\documentclass{article}

\usepackage[english]{babel}

\usepackage[letterpaper,top=2cm,bottom=2cm,left=3cm,right=3cm,marginparwidth=1.75cm]{geometry}

\usepackage{float}
\usepackage{amsmath}
\usepackage{amssymb}
\usepackage{graphicx,epstopdf}
\usepackage{subcaption}
\usepackage{authblk}
\usepackage{array}
\usepackage[colorlinks=true, allcolors=blue]{hyperref}
\usepackage{tabularx}

\title{Efficient and Flexible Method for Reducing Moderate-size Deep Neural Networks with Condensation}
\author[$^{1}$]{Tianyi Chen}
\author[$^{1}$]{Zhi-Qin John Xu \thanks{Corresponding author: xuzhiqin@sjtu.edu.cn}}
\affil[$^{1}$]{School of Mathematical Sciences, Institute of Natural Sciences, MOE-LSC, Shanghai Jiao Tong University}
\begin{document}
\maketitle

\begin{abstract}
Neural networks have been extensively applied to a variety of tasks, achieving astounding results. Applying neural networks in the scientific field is an important research direction that is gaining increasing attention. In scientific applications, the scale of neural networks is generally moderate size, mainly to ensure the speed of inference during application. Additionally, comparing neural networks to traditional algorithms in scientific applications is inevitable. These applications often require rapid computations, making the reduction in neural network sizes increasingly important. Existing work has found that the powerful capabilities of neural networks are primarily due to their nonlinearity. Theoretical work has discovered that under strong nonlinearity, neurons in the same layer tend to behave similarly, a phenomenon known as condensation. Condensation offers an opportunity to reduce the scale of neural networks to a smaller subnetwork with a similar performance. 
In this article, we propose a condensation reduction method to verify the feasibility of this idea in practical problems, thereby validating existing theories. 
Our reduction method can currently be applied to both fully connected networks and convolutional networks, achieving positive results. In complex combustion acceleration tasks, we reduced the size of the neural network to 41.7\% of its original scale while maintaining prediction accuracy. In the CIFAR10 image classification task, we reduced the network size to 11.5\% of the original scale, still maintaining a satisfactory validation accuracy. Our method can be applied to most trained neural networks, reducing computational pressure and improving inference speed.

\end{abstract}

\section{Introduction}

Neural networks have achieved globally astounding results, demonstrating an exceptional performance across a range of tasks in scientific fields including biology, medicine, astronomy, environmental science, physics, chemistry, etc.~\cite{reiser2022,Sarvamangala2022,shlomi2020,Smith2023, Zhang2021,Zhong2021}. Serving as a novel numerical solution tool, neural networks have been effectively applied in solving partial differential equations within various domains~\cite{Raissi2019,blechschmidt2021three,li2020neural,michoski2020solving}.  
Neural networks can learn multiscale models from data by designing appropriate network structures and sampling methods, offering insights into these intricate systems~\cite{xu2024solving}.

Currently, the scale of neural networks applied in the scientific field is generally moderate size~\cite{Raissi2019}. Compared to the industrial and internet sectors, obtaining experimental data in the scientific field often incurs high costs, resulting in significantly fewer data samples available for neural network training~\cite{Raissi2019}. This limitation in sample size restricts the scale of networks that can be trained. In addition, training large-scale neural networks requires expensive computational resources. The computing conditions in academia generally lag behind those in the industrial sector, making it challenging to support the training and inference of overly large neural network models.

Reducing the scale of neural networks applied in the scientific field holds significant importance~\cite{liu2019rethinking,wang2021convolutional}. Many traditional numerical algorithms, such as multigrid methods and fast Fourier transforms, exhibit quasi-linear or even linear time complexity, boasting an extremely high computational efficiency~\cite{wesseling1995introduction}. When applying neural networks to time-sensitive scientific computing tasks, it is crucial to reduce the network size to enhance its inference speed~\cite{xu2024solving}. Additionally, many scientific applications require deploying models in resource-constrained environments, such as embedded systems, mobile devices, and sensors~\cite{erichson2020shallow,Resource2024}. The computational power of these deployment platforms often falls far short of the platforms used for the original training of the tasks. Therefore, it is essential to reduce the network size as much as possible while maintaining the performance of the original network model, enabling efficient deployment in these contexts.

Nonlinearity is the fundamental characteristic that endows neural networks with their powerful expressive, learning, and generative capacities~\cite{Phase2021}. Nonlinear activation functions grant neural networks the ability to learn complex patterns, enabling neurons to model complex nonlinear relationships and approximate any continuous function with arbitrary precision. The superposition of multiple layers of nonlinear transformations allows neural networks to construct highly nonlinear decision boundaries and learn complex nonlinear mappings between inputs and outputs. Theoretical findings indicate that in the presence of strong nonlinearity within neural networks, neurons in the same layer tend to exhibit a condensation phenomenon, known as condensation~\cite{Phase2021}. Condensation refers to the gradual alignment of the direction of neurons' parameter vectors during the training process, implying a uniform preference for the inputs from the previous layer. This phenomenon is prevalent across various neural network architectures, suggesting that condensed neurons function similarly, or are nearly equivalent to a single neuron. Condensation implies the presence of a large number of similar neurons in neural networks, indicating that the structural complexity of neural networks may not be as high as it appears. When the network reaches an extreme point, the condensation phenomenon becomes more pronounced. In this case, the embedding principle suggests that the extreme point reached by the network is actually an extreme point of one of its subnetworks~\cite{zhang2022embedding}. 

Based on the above idea, we propose the condensation reduction method to validate existing theories.
The main contribution of this paper is the successful validation of the universality of the condensation phenomenon and the applicability of the embedding principle in practical applications through the condensation reduction method.
The use of the condensation reduction method requires a pretrained network, which could be considered a drawback. However, this limitation might be quite common, as seen in the work related to the ``Lottery Ticket Hypothesis''~\cite{frankle2018lottery}.

From the perspective of reduction algorithms, our method can be approximated as a pruning method. The key difference is that our approach merges branches instead of merely removing redundant ones. The work by Liu et al.~\cite{liu2021using} proposed a network condensation method, but their definition of condensation differs from ours. 
In their work, the term ``condense'' is primarily used as a verb to denote operations such as simplifying, refining, and compressing the network. However, in this study, ``condensation'' is defined as a significant phenomenon commonly observed during neural network training and can be directly observed using cosine similarity.
Most pruning methods measure the importance of parameters based on certain metrics and remove the less important ones~\cite{cheng2023survey}, which can be either structured or unstructured. These metrics usually include parameter norms, impact on loss, and sensitivity~\cite{hanson1988comparing,lecun1989optimal,you2019gate}.

In contrast, our condensation reduction method is a structured network reduction approach. Unlike structured pruning, we do not delete neurons or parameters based on metrics. Instead, we merge neurons based on their similarity (degree of condensation). Merging neurons involves deleting redundant neurons and modifying the parameters of the retained neurons, ensuring that the merged neurons have similar expressive capabilities as before. In the following sections, we prove the consistency of the neural network's output before and after the merger under strict conditions.


By leveraging the condensation phenomenon for neural network reduction, we aim to strike a balance between model performance and model size to identify an appropriate subnet. Utilizing the embedding principle, we understand that when a neural network reaches an extremum, it is likely at an extremum of a smaller subnet, making the neural network approximately equivalent to this subnet~\cite{zhang2022embedding}. At this point, we merge neurons that have condensed into a single neuron, resulting in a subnet of the original network and thereby achieving a reduction in the neural network's scale. 
The reduction based on the condensation phenomenon requires only the calculation of angles between neurons, making the time required for each reduction negligible compared to the training time. 
Owing to the universality of the condensation phenomenon, our reduction algorithm can be broadly applied to various types of models.

We are currently capable of reducing two major categories of neural networks: fully connected neural networks (FNNs) and convolutional neural networks (CNNs), achieving promising results. We selected two representative tasks for model reduction: fitting and classification. The fitting task involved neural network acceleration for combustion simulation, while the classification task focused on image classification on the CIFAR10 dataset, corresponding to fully connected neural networks and convolutional neural networks, respectively. In the acceleration of the combustion simulation, we reduced the original model to 40.9\% of its parameter count, maintaining consistency in both zero-dimensional and one-dimensional combustion simulations with the original model, and observed virtually no difference in complex turbulent flame simulations. For the CIFAR10 image classification task, we reduced the model to 11.5\% of its original parameter count, with the classification accuracy dropping only to 94\% of the original accuracy.

The rest of the article is organized as follows: The second part, Materials and Methods, introduces the concept of condensation and the details of condensation reduction. The third part, Results, presents the outcomes of condensation reduction in combustion tasks and CIFAR10 image classification tasks. The fourth part, Discussion, provides a discussion on the results and methods.

\section{Materials and Methods}
\subsection{Condensation reduction of FNN}
\subsection{FNN}
Considering an $(L+1)~$-layer fully connected neural network with a structure of

\begin{equation}
d_{in}-m_1-...-m_{i}-...-m_L-d_{out},
\end{equation}
where $m_i$ is the width of the $i$-th hidden layer.
The neural network can be defined as

\begin{equation}
\begin{gathered}
\boldsymbol{\vec{x}}_{[0]}=(\boldsymbol{\vec{x}}, 1), \quad \boldsymbol{\vec{x}}_{[i]}=\left(\sigma\left(\boldsymbol{W}_{[i]} \boldsymbol{\vec{x}}_{[i-1]}\right), 1\right), \text { for } i \in\{2,3, \ldots, L\} \\
f(\boldsymbol{\theta}, \boldsymbol{x})=\boldsymbol{a}^{\boldsymbol{\top}} \boldsymbol{\vec{x}}_{[L]} \triangleq f_{\boldsymbol{\theta}}(\boldsymbol{\vec{x}}),
\end{gathered}
\end{equation}
where $\boldsymbol{W}_{[i]} = (W_{[i]}^0,\boldsymbol{\vec{b}}_{[i]}^{\boldsymbol{\top}})$ consists of both weight matrix and bias vector, and $\sigma(\cdot)$ is the activation function. 
The parameter matrix of the $i$-th layer of the neural network is

\begin{equation}
W_{[i]}^{m_i\times m_{i-1{}}} = \begin{bmatrix}
\boldsymbol{\vec{v}}_1 \\
\boldsymbol{\vec{v}}_2 \\
\vdots \\
\boldsymbol{\vec{v}}_{m_i}
\end{bmatrix},
\end{equation}
wherein $\boldsymbol{\vec{v}}_t$ is the $t$-th neuron in the $i$-th layer.

\subsubsection{Cosine similarity}

We define the Cosine similarity between the $i$-th layer neurons as

\begin{equation}
S(\boldsymbol{\vec{v}}_i, \boldsymbol{\vec{v}}_j)=\frac{\boldsymbol{\vec{v}}_i \boldsymbol{\vec{v}}_j^{\boldsymbol{\top}}}{\left(\boldsymbol{\vec{v}}_i \boldsymbol{\vec{v}}_i^{\boldsymbol{\top}}\right)^{1 / 2}\left(\boldsymbol{\vec{v}}_j \boldsymbol{\vec{v}}_j^{\boldsymbol{\top}}\right)^{1 / 2}}.
\end{equation}
When considering neurons as vectors, if the cosine similarity between them is very close to 1, their directions are very similar, indicating that these two neurons are highly similar. The calculation of the cosine similarity is straightforward. Furthermore, in the following section, we will demonstrate that when the cosine similarity is exactly $1$, the functions expressed by the original network and the reduced network are consistent.

\subsubsection{Condensation of FNN}
Considering a neural network with a single hidden layer, whose activation function satisfies homogeneity, i.e. satisfies

\begin{equation}
\begin{aligned}
\sigma(k\cdot input) &= k\sigma(input),~~ \forall k \in \mathbb{R}^+ \\
f_{\boldsymbol{\theta}}(\boldsymbol{x}) &=\sum_{i=1}^m a_i \sigma\left(\boldsymbol{\vec{v}}_i \cdot \boldsymbol{x}\right).
\end{aligned}
\end{equation}
ReLU activation function is an activation function that satisfies homogeneity.

If the directions of the $p$-th and $q$-th neurons are the same, i.e, when the cosine similarity is $1$, we have
\begin{equation}
\boldsymbol{\vec{v}}_q = \lambda \boldsymbol{\vec{v}}_p, ~\lambda \in \mathbb{R}^+.
\end{equation}
The neural network at this point can be represented as
\begin{equation}
\begin{aligned}
f_{\boldsymbol{\theta}}(\boldsymbol{x}) & =\sum_{i=1}^m a_i \sigma\left(\boldsymbol{\vec{v}}_i \cdot \boldsymbol{x}\right) \\
& = \sum_{i\neq p,q}^m a_i \sigma\left(\boldsymbol{\vec{v}}_i \cdot \boldsymbol{x}\right) + a_p\sigma\left(\boldsymbol{\vec{v}}_p \cdot \boldsymbol{x}\right) + a_q\sigma\left(\boldsymbol{\vec{v}}_q \cdot \boldsymbol{x}\right)\\
& = \sum_{i\neq p,q}^m a_i \sigma\left(\boldsymbol{\vec{v}}_i \cdot \boldsymbol{x}\right) + (a_p+\lambda a_q)\sigma\left(\boldsymbol{\vec{v}}_p \cdot \boldsymbol{x}\right)\\
& \triangleq  f^{[new]}_{\boldsymbol{\theta}}(\boldsymbol{x}).
\end{aligned}
\end{equation}
At this point, the function expression of the neural network is completely identical to the new neural network formed by merging neurons $p$ and $q$. This means that neuron $p$ and neuron $q$ are equivalent to a single neuron at this moment. The above derivation can be generalized to any fully connected neural network of arbitrary depth that meets the criteria of homogeneity. 

\subsubsection{Condensation Reduction}
Leveraging the phenomenon of condensation for reduction involves merging neurons that have condensed into a single neuron. Determining whether two neurons have condensed is a matter of judging whether their directions are similar. Thus, we only need to artificially set a threshold for cosine similarity. When the cosine similarity between two neurons exceeds this threshold, they are considered to have condensed. We will henceforth refer to the cosine similarity threshold as the condensation threshold.

To perform condensation-based reduction, we first need to obtain the cosine similarity among neurons in the $l$-th layer targeted for reduction. By utilizing the formula for cosine similarity, we can easily calculate it using the parameter matrix of that layer of the neural network, specifically:
\begin{equation}
C_{[l]}=D_{[l]}A_{[l]}D_{[l]},
\end{equation}
wherein,
\begin{equation}
A_{[l]}=(a_{ij})=W_{[l]}W_{[l]}^{\boldsymbol{\top}}, 
~~~~~~~D_{[l]}=\text{diag}(\frac{1}{\sqrt{a_{11}}},...,\frac{1}{\sqrt{a_{m_lm_l}}}).
\end{equation}
After obtaining the cosine similarities among neurons, we need to group neurons that have condensed together. In each grouping session, given the condensation threshold, we calculate the number of neurons that have condensed with each ungrouped neuron. Subsequently, we select the neuron with the highest number of condensed neurons as the main neuron for that group.

After grouping all neurons, the next step is to merge the condensed neurons. We merge all neurons in each group towards the main neuron of that group. Let's assume, for simplicity, that the first $N$ neurons are classified into one group. Then, we have
\begin{equation}
\boldsymbol{\vec{u}}_{main}^{[new]} = \sum_k^{N}\frac{\|\boldsymbol{\vec{v}}_{k}\|_2}{\|\boldsymbol{\vec{v}}_{main}\|_2}\boldsymbol{\vec{u}}_{k},
~~~~~~~\boldsymbol{\vec{v}}_{main}^{[new]}=\boldsymbol{\vec{v}}_{main},
\end{equation}
wherein,
\begin{equation}
W_{[l]}^{m_i\times m_{i-1{}}} = \begin{bmatrix}
\boldsymbol{\vec{v}}_1 \\
\vdots \\
\boldsymbol{\vec{v}}_N \\
\boldsymbol{\vec{v}}_{N+1}\\
\vdots \\
\boldsymbol{\vec{v}}_{m_i}
\end{bmatrix},
~~~~~~~W_{[l+1]}^{m_{i+1}\times m_{i}} = \begin{bmatrix}

\boldsymbol{\vec{u}}_1^{\boldsymbol{\top}} &  \cdots & \boldsymbol{\vec{u}}_N^{\boldsymbol{\top}} & \boldsymbol{\vec{u}}_{N+1}^{\boldsymbol{\top}} & \cdots & \boldsymbol{\vec{u}}_{m_i}^{\boldsymbol{\top}}

\end{bmatrix}.
\end{equation}
The new neural network parameter matrix obtained is
\begin{equation}
W_{[l]_{new}}^{(m_i-N+1)\times m_{i-1{}}} = \begin{bmatrix}

\boldsymbol{\vec{v}}_{main}^{[new]} \\
\boldsymbol{\vec{v}}_{N+1}\\
\vdots \\
\boldsymbol{\vec{v}}_{m_i}
\end{bmatrix},
~~~~~~~W_{[l+1]_{new}}^{m_{i+1}\times (m_{i}-N+1)} = \begin{bmatrix}
 \boldsymbol{\vec{u}}_{main}^{{[new]}^{\boldsymbol{\top}}} & \boldsymbol{\vec{u}}_{N+1}^{\boldsymbol{\top}} & \cdots & \boldsymbol{\vec{u}}_{m_i}^{\boldsymbol{\top}}
\end{bmatrix}.
\end{equation}
By using the new neural network parameter matrix, we can create a new neural network, thus completing the condensation-based reduction process.

\subsection{Condensation reduction of CNN}
\subsubsection{CNN}
Convolutional neural networks can be understood as fully connected neural networks with shared weights. Therefore, all the definitions related to fully connected networks mentioned above can naturally be extended to apply to CNNs as well.

Consider a neural network that contains only convolutional layers and fully connected layers; its structure is as follows:
\begin{equation}
d_{in}-m_{C}^{[1]}-...-m_{C}^{[i]}-...-m_{C}^{[L]}-m_{F_1}-m_{F_2}-d_{out},
\end{equation}
wherein,
\begin{equation}
m_{C}^{[i]} = (C_{inchannel}^{[i]},C_{outchannel}^{[i]},\text{Size}^{[i]});
~~~~~~~\text{Size}^{[i]}=s^{[i]}\times s^{[i]}.
\end{equation}
wherein $C_{inchannel}^{[i]}$ and $C_{outchannel}^{[i]}$ are the dimension of input channel and output channel of $i$-th layer, $\text{Size}^{[i]}$ is the size of convolution kernel.
We view the $i$-th convolutional layer as containing outchannel neurons, each neuron being a convolutional kernel, represented by a vector as follows:
\begin{equation}
\boldsymbol{\vec{v}}_i^{C_{inchannel}^{[l]}\times \text{Size}^{[l]}} =(k_1^1,\cdots,k_{\text{Size}^{[l]}}^1,\cdots,k_1^{C_{inchannel}^{[l]}},\cdots,k_{\text{Size}^{[l]}}^{C_{inchannel}^{[l]}}).
\end{equation}
The parameter matrix of this layer can be represented as
\begin{equation}
W_{[l]}^{C_{outchannel}^{[l]}\times (C_{inchannel}^{[l]}\times \text{Size}^{[l]})} = \begin{bmatrix}
\boldsymbol{\vec{v}}_1 \\
\boldsymbol{\vec{v}}_2 \\
\vdots \\
\boldsymbol{\vec{v}}_{C_{outchannel}^{[l]}}
\end{bmatrix}.
\end{equation}
Corresponding to the small data processed by each convolution kernel, a layer of convolution operations can be represented as
\begin{equation}
y^{C_{outchannel}^{[l]}} = \sigma(W_{[l]}^{C_{outchannel}^{[l]}\times (C_{inchannel}^{[l]}\times \text{Size}^{[l]})} \cdot x^{C_{inchannel}^{[l]}\times \text{Size}^{[l]}})^{\boldsymbol{\top}},
\end{equation}
Then we calculate the cosine similarity between neurons $\boldsymbol{\vec{v}}_i$ according to the previous definition.

\subsubsection{Condensation of CNN}
Similar to fully connected neural networks, when there are two neurons in a convolutional layer with the same orientation, the function represented by the network is completely identical to the function represented by the new network after merging the two neurons. Refer to the appendix~\ref{A_condensation_cnn} for proof details.
Therefore, for convolutional neural networks, the phenomenon of condensation will be consistent with that of fully connected networks, thereby reducing the complexity of the model.

\subsubsection{Condensation reduction}
The traditional definition of condensation reduction in convolutional neural networks, following the above derivation, is completely identical to that in fully connected networks. It simply requires unfolding each convolutional layer into a matrix composed of convolutional kernel parameters, specifically:
\begin{equation}
W_{[l]}^{C_{outchannel}^{[l]}\times (C_{inchannel}^{[l]}\times \text{Size}^{[l]})} = \begin{bmatrix}
\boldsymbol{\vec{v}}_1 \\
\boldsymbol{\vec{v}}_2 \\
\vdots \\
\boldsymbol{\vec{v}}_{C_{outchannel}^{[l]}}
\end{bmatrix}.
\end{equation}
Thus, the corresponding cosine similarity matrix can be calculated according to the previous definition.
Afterward, the neurons that have condensed are grouped in the exact same manner as during the condensation reduction of fully connected networks. Subsequently, neurons belonging to the same group are merged into a single neuron.

When the layer following the condensation-reduced layer is a fully connected layer, the process of obtaining new network parameters is entirely the same as during the condensation reduction of fully connected networks. When the layer following the condensation-reduced layer is a convolutional layer, the situation is, in fact, consistent with that of fully connected networks, details are shown in the appendix~\ref{A_detials_reduc_cnn}.

The concept of condensation reduction can indeed be extended to various types of convolutional layers, such as depthwise separable convolutions, achieving satisfactory application results. Taking the structure of MobileNetV2 as an example, a complete module within MobileNetV2 consists of a sequence that includes a pointwise convolution followed by batch normalization and a ReLU6 activation function; a depthwise convolution followed by batch normalization and a ReLU6 activation function; and another pointwise convolution followed by batch normalization and a linear activation function~\cite{sandler2019mobilenetv2}. In such a complete module, the depthwise convolution layer plays a primary role, while the pointwise convolution layers mainly serve to expand and reduce dimensions. Therefore, condensation reduction is applied only to the depthwise layer. Due to the characteristics of the complete module, reducing the depthwise layer accordingly reduces the size of the adjacent pointwise layers. The details of derivation for the depthwise separable convolutions are shown in the appendix~\ref{A_detials_reduc_dwcnn}.

\subsection{Manual condensation reduction and automatic condensation reduction}
\subsubsection{Manual condensation reduction}
In the face of complex application scenarios, neural networks can present challenging metrics that are intricate, time-consuming, or require subjective judgment by supervisors. In such cases, it is necessary to pause and conduct model testing each time a reduced model is obtained. Manual condensation reduction is primarily employed in these situations.

Manual condensation reduction involves reducing a neural network layer by layer after achieving a neural network that meets performance standards. This approach ensures that the neural network does not deviate significantly due to a single reduction. Each manual condensation reduction requires setting a condensation threshold. The appropriate threshold is determined by observing changes in the loss before and after reduction. If the loss increases significantly after reduction, the condensation threshold is raised; if the loss is too small, the threshold is lowered. If the neural network's performance meets the requirements after a certain amount of training post-reduction, further reduction is performed; otherwise, the condensation threshold is increased, and the process is repeated. This cycle continues until an appropriately sized subnet is obtained. An appropriate subnet must meet two criteria: first, it must achieve the set performance metrics, such as accuracy in combustion simulation predictions. Secondly, the neural network parameters should be difficult to reduce further, indicating that the number of neurons condensing together is scarce or that even slight reductions significantly impact the network's performance.

\subsubsection{Automatic condensation reduction}
\begin{figure}[H]
\includegraphics[width=13.5 cm]{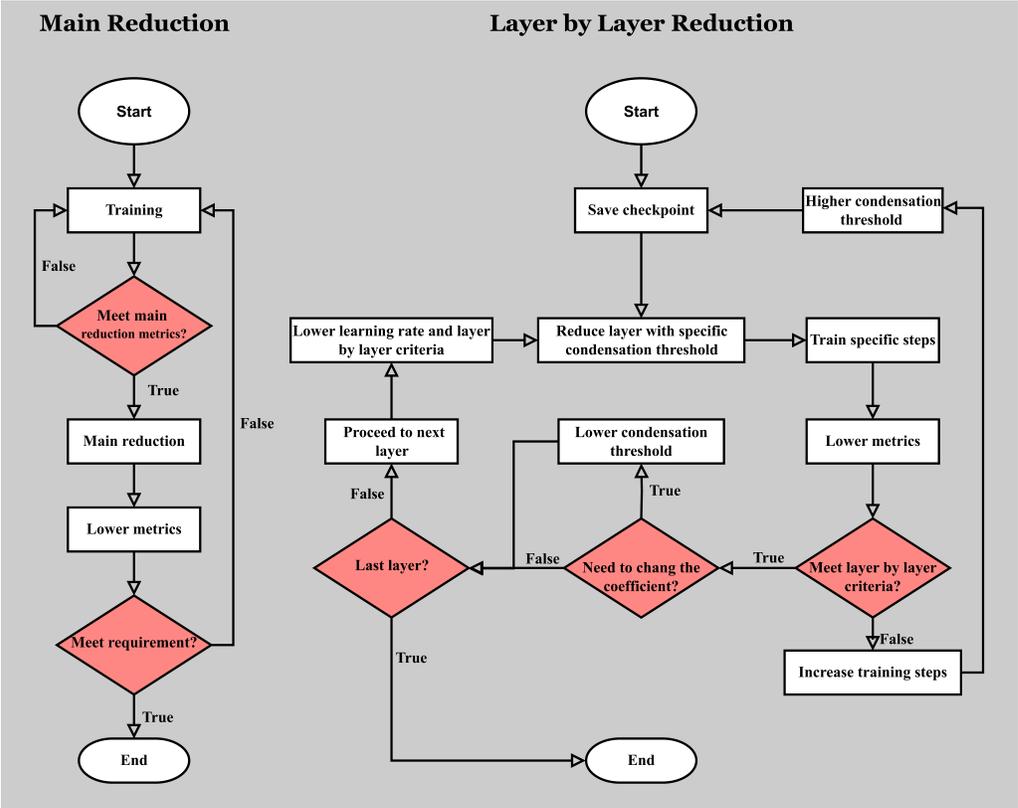}
\caption{Flow chart of automatic condensation reduction.The left side shows the main reduction process of automatic condensation reduction, while the right side details the layer-by-layer reduction process carried out in each main reduction.}
\label{flow_1}
\end{figure}

The main steps of automatic condensation reduction are consistent with manual condensation reduction, except that the reduction process does not require manual control. Automatic condensation reduction is mainly used in scenarios with clear neural network performance metrics and where testing is convenient and rapid, such as using validation accuracy to assess performance in image classification problems. The process of automatic condensation reduction is shown in the figure~\ref{flow_1}.

\unskip

The main flow of automatic condensation reduction begins with training the original model and periodically testing its performance. Once the test meets the main reduction criteria, main reduction is carried out. After the main reduction, the reduced model continues training until it meets the main reduction criteria again, and this cycle repeats until the model reduction reaches the anticipated extent. After each main reduction, the criteria for the next main reduction are slightly lowered because reduction inevitably affects the network's performance. Keeping the same criteria would make the reduction cycle very slow or even impossible, wasting a lot of time on model training.

Main reduction involves sequentially reducing each layer of the model. Before each layer-by-layer reduction, the model's relevant data are saved, creating a save point, followed by the reduction. For each layer to be reduced, a specific condensation threshold is applied during its reduction. After reducing a layer, a specific number of training steps are performed, the same for each layer. Upon completion, the model is automatically tested to see if it meets the layer-by-layer reduction criteria, which are generally lower than the main reduction criteria to expedite the overall main reduction process. If the model meets the layer-by-layer criteria and the current layer is not the last to be reduced, reduction proceeds to the next layer. Simultaneously, the layer-by-layer criteria are lowered, as is the initial learning rate for the next reduction. Specifically, if the model's parameters have hardly decreased after a reduction, the specific layer's condensation threshold is lowered for a more aggressive reduction next time. If the model does not meet the layer-by-layer criteria, it reverts to the previous save point to restart reduction. Simultaneously, the specific training steps for layer-by-layer reduction increase, and the specific condensation threshold for that layer is raised. This aims to improve model performance by extending training time for the reduced model to recover performance and by stricter condensation reduction, as a higher condensation threshold means neurons deemed for condensation are more directionally aligned, thus reducing error upon merging. Since the condensation threshold can be adjusted up to 1, where the model remains unchanged, such layer-by-layer reduction can always be completed. When the last layer's reduction is finished, the main reduction concludes.

Automatic condensation reduction merely requires setting initial values for the algorithm beforehand. Aside from the main and layer-by-layer reduction criteria, the quality of other initial values does not sensitively affect the final reduction outcome, as the key condensation threshold can be automatically adjusted. However, excessively deviated initial values will waste considerable time on model training. Before using the automatic reduction algorithm, it is advisable to make reasonable estimates based on the training process of the model to be reduced. The main reduction criteria are mainly based on the highest performance of the model to be reduced and the minimum performance requirements of the reduced model. The automatic reduction algorithm ensures these criteria are met, obtaining a sub-network that is appropriately sized between the minimal sub-network that retains structure and the original network.

\section{Results}
\subsection{Acceleration for combustion simulation}
\subsubsection{Background on task}
Numerical simulation is indispensable in both scientific research and industrial production. In problems involving various multiscale dynamic systems, such as combustion, numerical simulation requires solving high-dimensional stiff ordinary differential equations, which traditional numerical solution methods often require very small time steps for~\cite{xu2024solving}.  Moreover, in numerical simulations of combustion, the direct integration of chemical reactions required by traditional methods is time-consuming~\cite{xu2024solving}. It has been proven reasonable and efficient to use neural networks to replace traditional numerical solution methods. Neural networks, by fitting state mapping functions, can map the current state of a system to the state of the system at the next time step. This method can be applied with larger time steps and replaces the time-consuming chemical reaction calculation process with the rapid inference process of neural networks~\cite{xu2024solving}.

The simulation of combustion requires integrating computational fluid dynamics (CFD) with chemical reactions. In this task, a fully connected neural network computes the rate of change of chemical substances as source terms for the CFD. The CFD code used is EBI~\cite{zirwes2020quasi}, and the chemical mechanism employed is the drm19 mechanism for methane, which involves 21 components and 84 reactions. In the depicted turbulent ignition test, as shown in figure~\ref{FNN_Tur2D}, a computational domain of 1.5cm × 1.5cm is set with 512 × 512 cells. The velocity field is generated using the Passot-Pouquet isotropic kinetic energy spectrum. The initial conditions are set at $T=300K$, $P=1atm$, and $\phi=1$. An ignition round is placed at the center of the domain, with a radius of 0.4mm. The figure~\ref{FNN_Tur2D} compares the results at 1ms simulated using the fully connected neural network and CVODE, both utilizing EBI. For more details, please refer to the original text~\cite{xu2024solving}.

The task at hand involves condensing and simplifying the classic neural network fitting problem applied in a real-world scenario using fully connected neural networks. It aims to validate the occurrence of traditional defined condensation phenomena in practical applications. Specifically, due to the complexity of combustion simulation, it can highlight the ability of condensation reduction to maintain the performance of neural networks.

\subsubsection{Training Setup}
\begin{figure}[h]
    \centering
    \begin{subfigure}[b]{0.45\textwidth}
        \includegraphics[width=\textwidth]{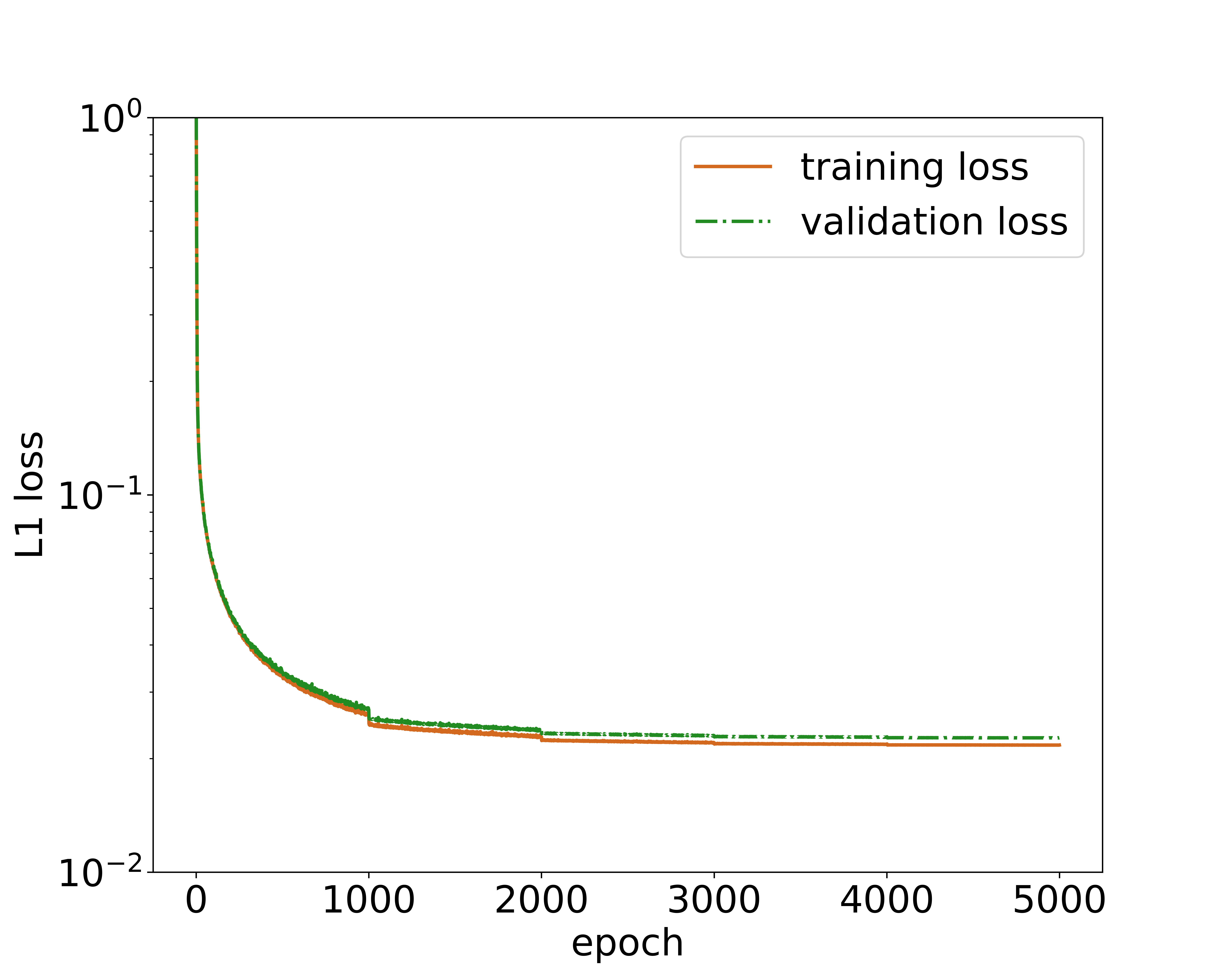}
        \caption{Original FNN model.}
        \label{FNN_LOSS_0}
    \end{subfigure}
    \begin{subfigure}[b]{0.45\textwidth}
        \includegraphics[width=\textwidth]{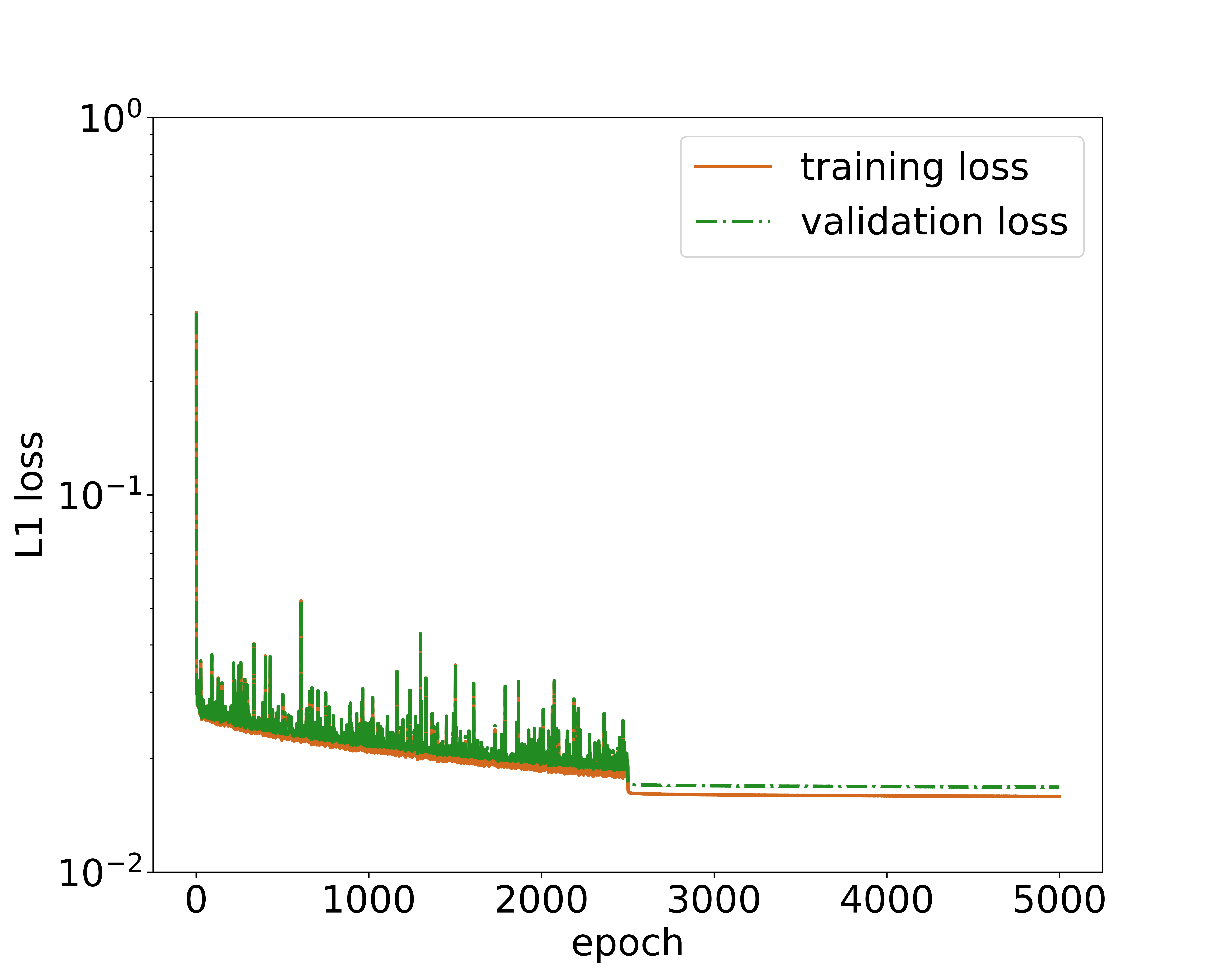}

        \caption{First FNN reduction model.}
        \label{FNN_LOSS_1}
    \end{subfigure}

    \vspace{0.419cm}

    \begin{subfigure}[b]{0.45\textwidth}
        \includegraphics[width=\textwidth]{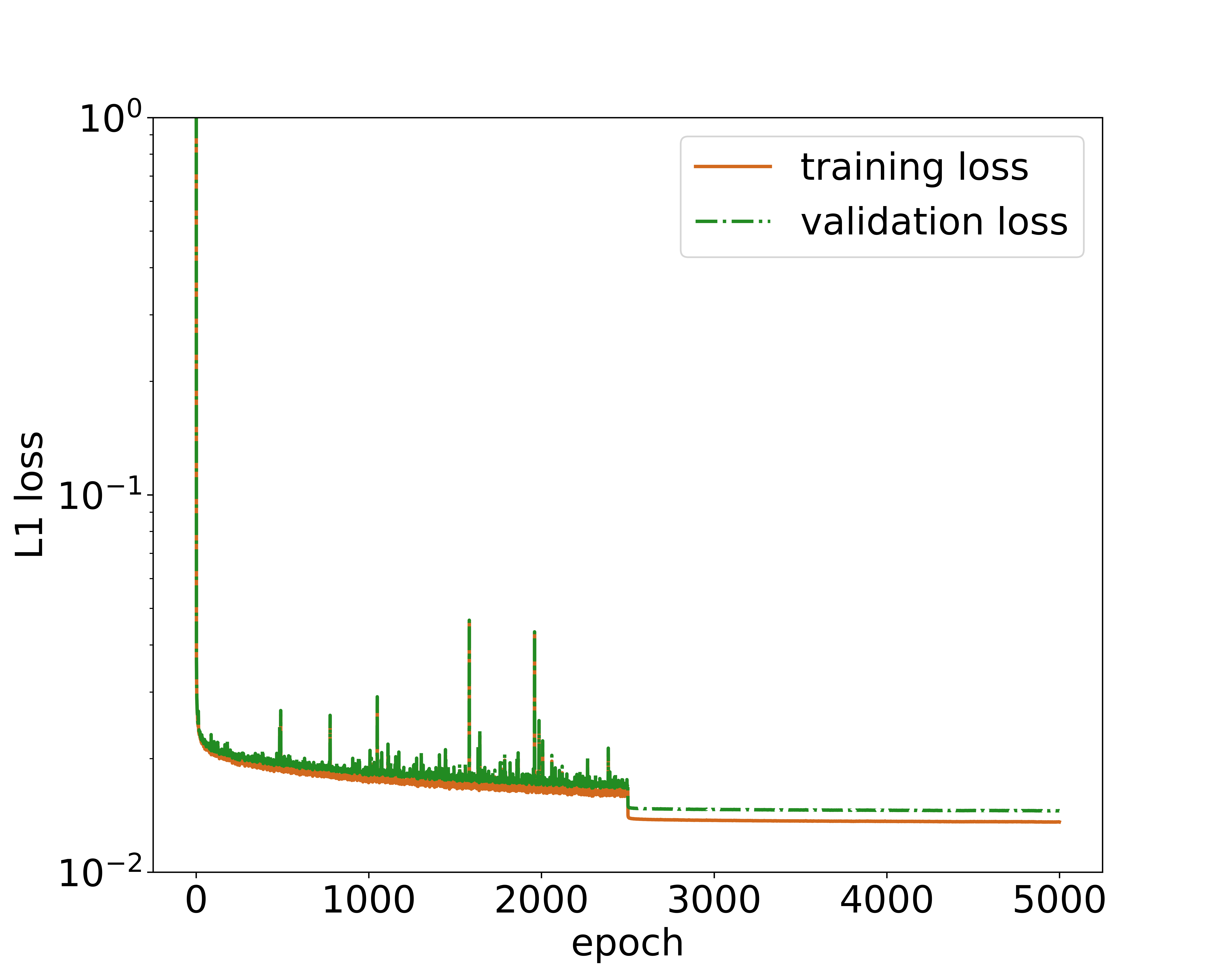}
        \caption{Second FNN reduction model.}
        \label{FNN_LOSS_2}
    \end{subfigure}
    \begin{subfigure}[b]{0.45\textwidth}
        \includegraphics[width=\textwidth]{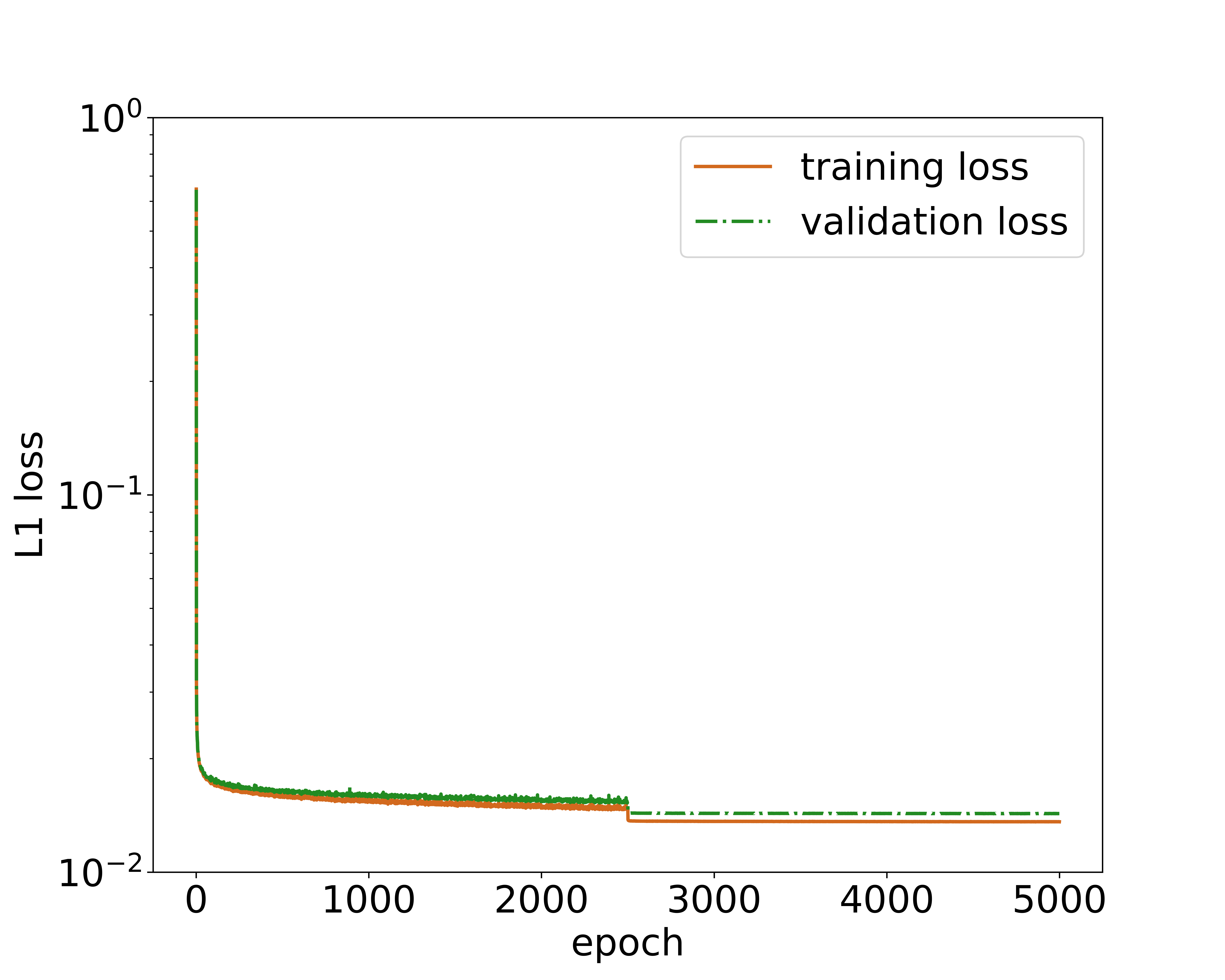}
        \caption{Third FNN reduction model.}
        \label{FNN_LOSS_3}
    \end{subfigure}

    \caption{Training loss and validation loss of the FNN models. The subfigures (\textbf{a}) to (\textbf{d}) illustrate the training and validation losses of the original fully connected neural network model through to its third reduction.}
    \label{fig:grid}
\end{figure}

The original model in this experiment is a fully connected neural network with the architecture of 23-3200-1600-800-400-23. The parameters of the neural network are initialized using the following Gaussian distribution.
\begin{equation}
N(0,var),var=(\frac{m_{in}+m_{out}}{2})^{-2},
\end{equation}
wherein $m_{in}$ and $m_{in}$ are input dimension and output dimensions of the specific layer of the model.
The initial learning rate is set to 0.0001 and is reduced to 50\% of its original value every 1000 steps to prevent the learning rate from becoming too large. The entire training process uses the Adam optimizer, sets 
 $betas=(0.9, 0.999)$, $eps=1 \times 10^{-8}$, $weight~decay=0$. The starting batch size is set to 1024 and is increased by a factor of 3.07 every 1000 steps to gradually accelerate training and enhance the model's generalization performance. 
The initial model undergoes training for 5000 steps, resulting in a training loss of 0.02172. Post-training, the model's accuracy across the 0th, 1st, and 2nd dimension is evaluated. The training and validation loss of the original model is shown in Figure~\ref{fig:grid}a.

\subsubsection{Reduction Setup and Result}
Due to the complex and time-consuming nature of evaluating neural network metrics in the combustion simulation task, a manual reduction approach is adopted. 
Please refer to Table~\ref{Comparison_FNN} for a comparison of the reduction processes. 
In the first reduction, the first layer of the original network is reduced, with a set condensation threshold of 0.9, resulting in a modified model architecture of 23-2205-1600-800-400-23. The initial learning rate is set at 0.0001, decreasing to 10\% of its previous value every 2500 steps, with Adam as the optimizer. The starting batch size is 1024, increasing to 128 times its original size every 2500 steps. After training for 5000 steps, the training loss is 0.01592. This reduction in loss is considered indicative of the model's performance being between that of the original model trained for 5000 steps and the original model trained for 10000 steps. This suggests that consolidating condensed neurons has a minimal impact on the neural network's performance. Furthermore, it is observed from the loss graph that the loss of the reduced neural network rapidly decreases to a satisfactory level within a few training steps, similar to the case of perturbing the parameters of the original model. This indicates that the condensed neurons are approximately equivalent to a single neuron, a finding that is validated by subsequent reductions.The training and validation loss is shown in the figure~\ref{FNN_LOSS_1}.

The second reduction further reduces the first layer of the neural network, with a set condensation threshold of 0.8, resulting in a model architecture of 23-1105-1600-800-400-23. This reduction uses the same training settings as the first reduction. After training for 5000 steps, the training loss is 0.01358, showing continued decline. At this point, the first layer of the original model has been reduced to 34.5\% of its initial size, yet it still maintains accuracy in the 0th and 1st dimensions and performs exceptionally well in two-dimensional turbulent flame simulations. This reflects that a significant number of neurons in the neural network are condensed, and the condensed model is approximately equivalent to the subnet resulting from the merged condensed neurons, further proving the feasibility of using condensation for network reduction. These two reductions indicate that the extreme points reached after training the original model are likely equivalent to those of a significantly smaller subnet in the first layer. The training and validation loss is shown in the figure~\ref{FNN_LOSS_2}.

\begin{figure}[H]
\includegraphics[width=13 cm]{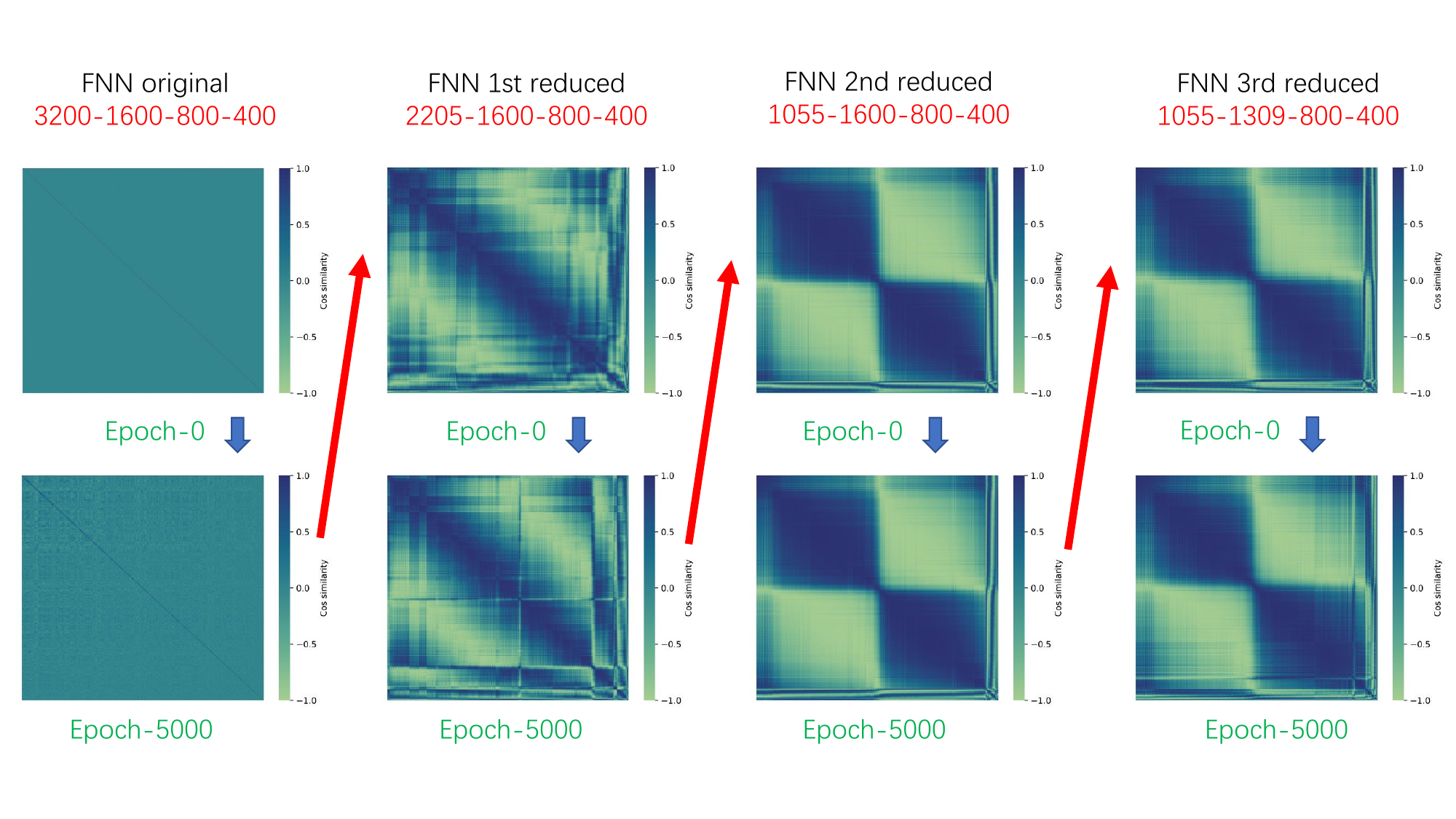}
\caption{Cosine similarity matrixs of 2nd layer on different stages. In the matrix heatmap, the element in the $i$-th row and $j$-th column represents the cosine similarity between the $i$-th neuron and the $j$-th neuron. The more distinct the blocks in the cosine matrix, the stronger the condensation of the neural network.
From left to right, the images correspond to the original fully connected neural network model through to the model after the third reduction. "Epoch 0" indicates that the model has just been reduced and has not yet been trained. "Epoch 5000" represents the fully connected neural network model upon completion of training.}
\label{flowcos}
\end{figure}

\begin{figure}[H]
\includegraphics[width=13 cm]{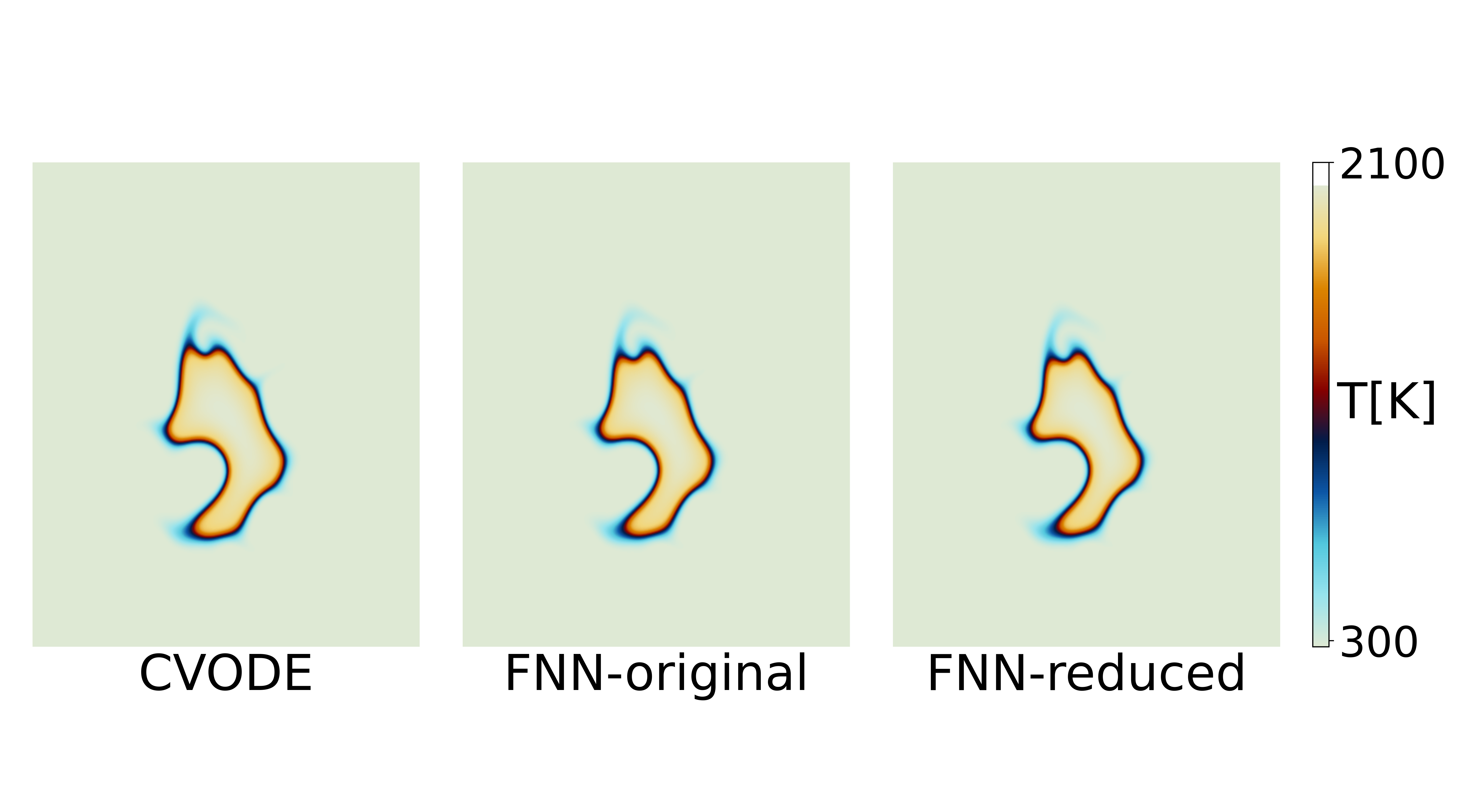}
\caption{Turbulent ignition. The image displays the temperature distributions obtained in combination with EBI for drm19, showing results from left to right for CVODE, the original fully connected neural network, and the neural network after the third reduction.}
\label{FNN_Tur2D}
\end{figure}

The third reduction targeted the second layer of the network, setting a condensation threshold of 0.999, resulting in a model architecture of 23-1105-1309-800-400-23. The condensation threshold was set so high because, after the first two reductions, the second layer of the neural network became highly condensed. A slightly lower condensation threshold would have led to a significant reduction in the network, potentially reducing the second layer to fewer than ten neurons and greatly impacting model performance. The phenomenon where reducing one layer of the neural network causes a higher degree of condensation in the subsequent layer is interesting. In practice, the network parameter matrices for the first and second layers, which we use for reduction, can be understood as the input and output parameter matrices of the first layer of the neural network, respectively. They are inherently related, and reducing the first layer of the network decreases the number of rows in the input parameter matrix while also decreasing the number of columns in the output parameter matrix. The relationship between them will be explored further in future research. As shown in the figure~\ref{flowcos}, it can be clearly seen that the reduction of the first layer of the neural network can immediately change the condensation state of the second layer of the neural network. After the first and second reductions, when examining the cosine similarity matrix of the second layer of the neural network in the untrained state, we can see that the previously inconspicuous condensation in the second layer appeared after the first reduction, and a strong condensation phenomenon emerged after the second reduction, making the second layer neural network capable of extremely strong condensation even at a high condensation threshold of 0.9. Since the previous model had already been trained for 15000 steps and reached a vicinity of a good extreme point, a smaller learning rate was used for this training phase. The initial learning rate was adjusted to 4e-5, with all other settings remaining the same as in the first reduction. After training for 5000 steps, the training loss was recorded at 0.01360, showing almost no change from the second reduction.The training and validation loss is shown in the figure~\ref{FNN_LOSS_3}. The results of the two-dimensional turbulence simulation of the reduced model are shown in the figure~\ref{FNN_Tur2D}, which is almost identical to the original model.
According to existing theories, we know that training a large network and gradually reducing its size can yield better results compared to directly training a small network. We retrained a small network with the same size as the reduced network, using the same training settings as previously described. The directly trained small network even failed to converge when predicting two-dimensional turbulent flames.

\unskip

\unskip

\begin{table}[H]
\caption{Comparison table of reduction process of FNN. ``Parameter'' represents the number of parameters in the neural network. Reduction rate represents the ratio of the current network parameter count to the original network parameter count. A lower reduction rate indicates a smaller network relative to the original, demonstrating a more effective reduction.}
\label{Comparison_FNN}
\begin{tabular}{lllll}
\hline
Model             & Structure               & Validation Loss & Parameter & reduction ratio \\ \hline
Original    & 23-3200-1600-800-400-23 & 0.02276         & 6,802,800 & 100\%                        \\ \hline
1st reduced & 23-2205-1600-800-400-23 & 0.01694         & 5,187,915 & 76.26\%                      \\ \hline
2nd reduced & 23-1105-1600-800-400-23 & 0.01454         & 3,402,615 & 50.01\%                      \\ \hline
3rd reduced & 23-1105-1309-800-400-23 & 0.01434         & 2,848,260 & 41.87\%                      \\ \hline
\end{tabular}
\end{table}

From the performance and training process of the model after three rounds of reduction, it is evident that each reduction preserved the performance capabilities of the previous model, demonstrating that the phenomenon of condensation ensures the quasi-equivalence between the original network and the subnetworks. The final reduced model has an architecture of 23-1105-1309-800-400-23, with a total of 2,848,260 parameters. In contrast, the original model, with an architecture of 23-3200-1600-800-400-23, had a total of 6,802,800 parameters, making the reduced model's size only 41.7\% of the original model. However, observations from various aspects show that the performance of the original and reduced models is nearly identical, proving the effectiveness of the reduction.

\subsection{Classification of CIFAR10}
\subsubsection{Background on task}
Image classification is a critical problem in neural network applications, and convolutional neural networks (CNNs) hold a prominent position among all neural network architectures. Choosing the image classification task can validate the universality of the condensation phenomenon in practical problems while demonstrating the wide applicability of condensation reduction.

The CIFAR10 dataset is a common dataset for image classification tasks, consisting of 60,000 32x32 color images categorized into 10 classes. Each class contains 6,000 images, totaling 50,000 training images and 10,000 testing images. Choosing the CIFAR10 dataset is due to the need for a reasonably sized network to achieve good performance and moderate training time, making it convenient for applying automatic condensation reduction method.

\begin{figure}[H]
\includegraphics[width=13 cm]{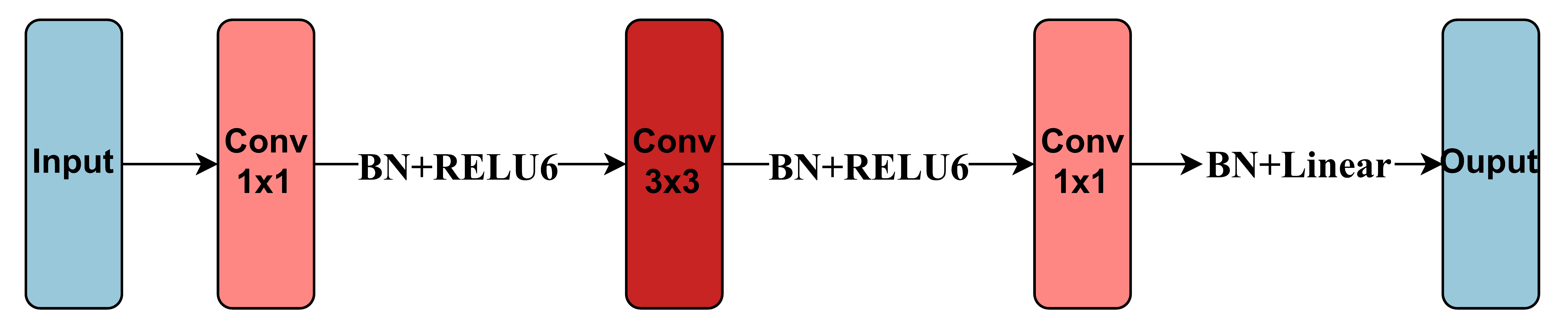}
\caption{Convolutional block of MobileNetV2. In the figure, 'Conv $1\times1$' refers to layers where the convolutional kernels are $1\times1$, with the number of channels and the number of kernels varying in each layer. 'Conv $3\times3$' indicates layers where the convolutional kernels are $3\times3$, with one channel, and the number of kernels may vary per layer. 'BN' stands for batch normalization, while 'ReLU6' and 'Linear' refer to activation functions.}
\label{block}
\end{figure}

MobileNetV2 is a convolutional neural network model that adopts depthwise separable convolutions and is dedicated to mobile device deployment~\cite{sandler2019mobilenetv2}. Its main network structure is shown in the tabel~\ref{Structure_Mobiel}. For detailed information about MobileNetV2, please refer to the original text~\cite{sandler2019mobilenetv2}. It concatenates 16 modules with the same structural characteristics, and the structure is shown in the figure~\ref{block}. The reasons for using MobileNetV2 for condensation reduction are as follows. First, MobileNetV2 targets mobile deployment scenarios, which is also an important application scenario for model reduction. MobileNetV2 itself contains a small number of parameters, and further reduction of MobileNetV2 can reflect the reduction capability and practical application value of the condensation reduction method. Second, MobileNetV2 has 16 modules with the same structural characteristics, which is very suitable for using automatic condensation reduction methods for layer-by-layer reduction. Finally, MobileNetV2 mainly adopts depthwise separable convolutions, which will further extend the definition of condensation and enrich the application fields of condensation reduction.
\begin{table}[H] 
\caption{Main structure of MobileNetV2~\cite{sandler2019mobilenetv2}.
'$1\times 1$' represents a convolutional kernel with a shape of $1\times 1$, where the number of channels is determined by the input dimensions. '$3\times 3$' refers to a convolutional kernel that is single-channel and has a shape of '$3\times 3$'. ReLU$6$ and linear are the activation functions used after the convolutional layers.The numbers in the table represent the count of convolutional kernels in each corresponding layer of the convolutional layer.
\label{Structure_Mobiel}}
\newcolumntype{C}{>{\centering\arraybackslash}X}
\begin{tabular}{llllllllllll}
\hline
Layer index          & 2  & 3   & 4   & 5-6 & 7   & 8-10 & 11  & 12-13 & 14  & 15-16 & 17  \\ \hline
$1\times 1$, ReLU$6$   & 96 & 144 & 144 & 192 & 192 & 384  & 384 & 576   & 576 & 960   & 960 \\ \hline
$3\times 3$, ReLU$6$ & 96 & 144 & 144 & 192 & 192 & 384  & 384 & 576   & 576 & 960   & 960 \\ \hline
$1\times 1$, linear  & 24 & 24  & 32  & 32  & 64  & 64   & 96  & 96    & 160 & 160   & 320 \\ \hline

\end{tabular}

\end{table}

\unskip

\subsubsection{Training and Reduction Setup}

For convolutional networks, we similarly adopt an initialization similar to that of fully connected networks, utilizing the following Gaussian distribution
\begin{equation}
N(0,var),var=(\frac{m_{in}+m_{out}}{2})^{-2},
\end{equation}
wherein $m_{in}$ and $m_{out}$ the are input dimension and output dimension of the specific layer of the model.

The initial learning rate is set to 0.01, and a cosine annealing strategy with a period length of 200 and a minimum learning rate of 0 is used for the first 400 steps. The optimizer during this phase is SGD with $momentum=0.9$. This part of the training can be seen as the process of quickly finding a good critical point for the model. After 400 steps, the learning rate changes to 0.001, and the optimizer switches to Adam, sets $betas=(0.9, 0.999)$, $eps=1 \times 10^{-8}$, $weight~decay=0$. Subsequently, every 1000 steps, the learning rate alternates between 0.001 and 0.0001. The main idea behind this learning rate strategy is to first use the cosine annealing strategy to find a suitable parameter region for the neural network and then use the Adam optimizer to refine the network parameters.

In our multiple experiments, we observed a noticeable improvement in test accuracy after switching to Adam. The accuracy obtained using this mixed strategy was higher compared to using Adam throughout the entire training process. Throughout the entire training process, the batch size remained at 128.

After each reduction, the above learning rate strategy will be reapplied for model training, but with a change in the initial learning rate. The method for changing the initial learning rate is
\begin{equation}
lr_{[new]} = lr_{min} + 0.5\times(lr_{max}-lr_{min})\times(1+\cos(\pi\times \frac{t_{reduction}}{T_{lr}})),
\end{equation}
wherein $t_{reduction}$ represents the current time of reductions, $T_{lr}$ is set to 200, $lr_{min}$ is set to 0.0001 and $lr_{max}$  is set to 0.01.
The reason for this approach is that each reduction can be considered as the model reaching a critical point that meets our requirements. At this point, the learning rate should be appropriately lowered to further enhance the model's performance. 
The model initiates main reduction process when the accuracy on the validation set reaches 0.88. Once the model achieves this main reduction criterion, it automatically begins layer-by-layer reduction. The criterion for layer-by-layer reduction is set at an accuracy of 0.84 on the validation set. After initiating layer-by-layer reduction, whenever the model meets this criterion, it automatically reduces the corresponding neural network layer. After each reduction, both the main reduction and layer-by-layer reduction criterion are automatically adjusted in the following manner:
\begin{equation}
Acc_{[new]}=
\left\{
\begin{aligned}
&Acc_{min} + 0.5\times(Acc_{max}-Acc_{min})\times(1+\cos(\pi\times \frac{t_{reduction}}{T_{Acc}})) &t_{reduction}<T_{Acc}\\
&Acc_{min} &t_{reduction}\geq T_{Acc},\\
\end{aligned}
\right.
\end{equation}
wherein $T_{Acc}$ is set to 100. $Acc_{max}$ and $Acc_{min}$ are set to 0.88 and 0.85 respectively for main reduction. $Acc_{max}$ and $Acc_{min}$ are set to 0.84 and 0.81 respectively for layer-by-layer reduction.
When the reduction is not the final layer in the layer-by-layer reduction, the initial limit for the number of training steps after reduction is set to 20 steps. If the layer-by-layer reduction criterion is not met within this limit, the model will revert to a previously saved checkpoint to re-attempt the reduction of that layer, with the limit on training steps permanently increasing by 10 steps. This limit on training steps is consistent for every layer except the final one in layer-by-layer reduction. This is because encountering difficulties in reducing one layer often implies that reducing each layer will pose a challenge.

The initial condensation threshold for each layer of the neural network is set at:
\begin{equation}
Cut=\frac{1}{1+\exp(-2)}.
\end{equation}
If a layer of the neural network fails to meet the layer-by-layer reduction criterion after reaching the limit of restricted training steps following reduction, the condensation threshold for that layer is set at:
\begin{equation}
Cut_{[new]}=\frac{1}{1+\exp(-2-0.1\times t_{fail})},
\end{equation}
wherein $t_{fail}$ represents the number of times reduction has failed.
If, after a reduction, the amount of parameters reduced in a certain layer of the neural network is too small, and the layer-by-layer reduction criterion can be met after reaching the limit of restricted training steps, then the condensation threshold for that layer is set at:
\begin{equation}
Cut_{[new]}=\frac{1}{1+\exp(-2-0.1\times t_{fail}^{[new]})},  ~~~~~~ t_{fail}^{[new]}=t_{fail}-1.
\end{equation}
A parameter reduction is considered too small if the proportion of the model's parameters after reduction, compared to before reduction, is greater than 0.999.

For the final layer in a layer-by-layer reduction, there are specific, larger limits for training steps and validation set accuracy criterion, as the reduction of the last layer often results in the largest decrease in parameters within the MobileNetV2 model, leading to significant changes in model performance. Here, the limit for training steps is set to 200, and the validation accuracy criterion is set at 0.8.

We have also established a criterion for significant model deviation following model reduction. If, after model reduction, the accuracy on the validation set is below 0.5 after 10 training steps, it is considered that the model has undergone a significant deviation, and it will be reverted to a previously saved checkpoint to undergo reduction again. This approach saves time that would have otherwise been spent training up to the limit of training steps.

Finally, upon achieving an appropriate parameter count through the main reduction cycle, the final classification layer is reduced using a condensation threshold of 0.4 for this last layer.

\subsubsection{Reduction Result}
\begin{figure}[H] 
    \centering
    \begin{subfigure}[b]{0.66\textwidth}
        \includegraphics[width=\textwidth]{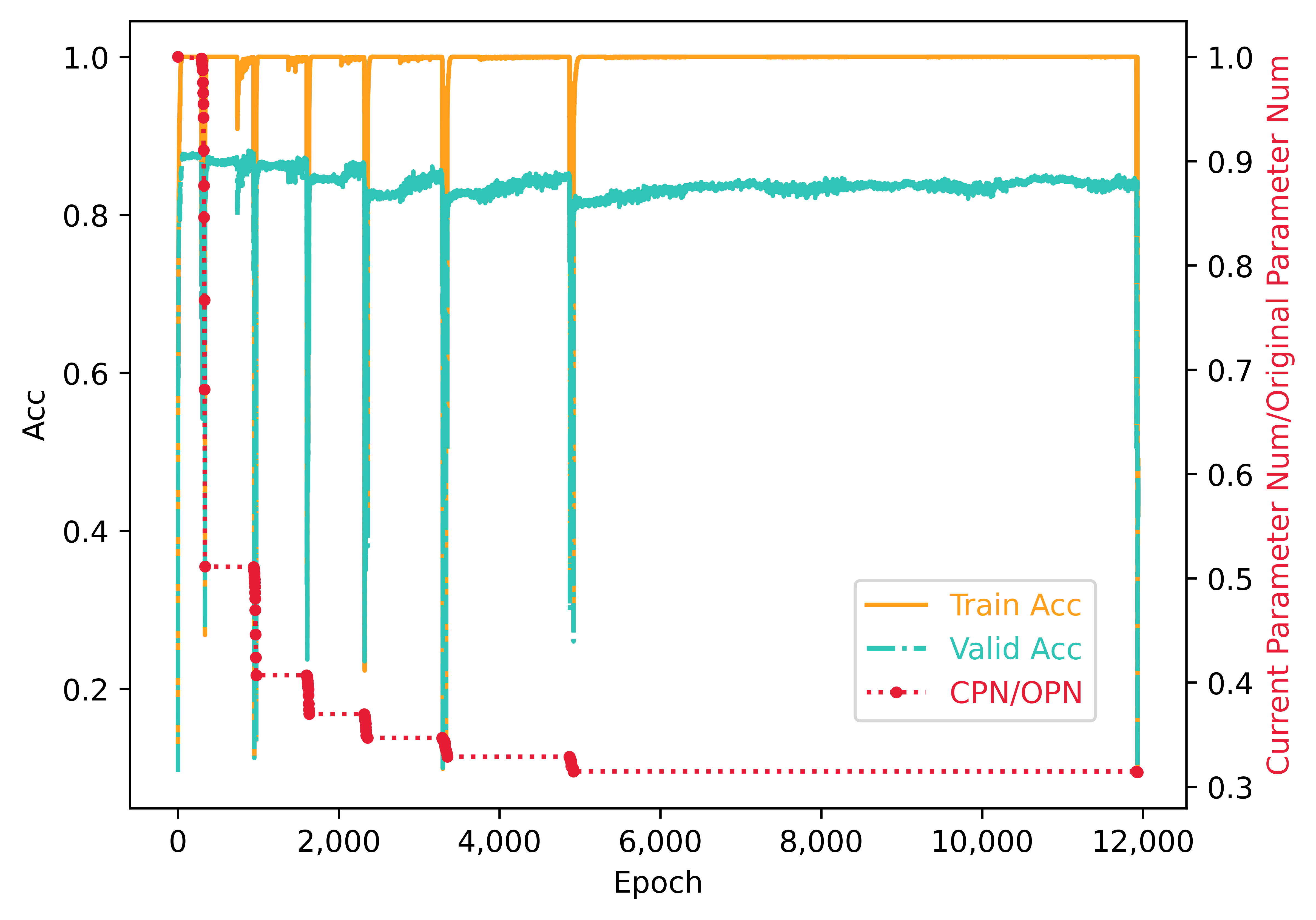}
        \caption{Accuracy.}
        \label{Conv_Acc}
    \end{subfigure}
    \vspace{0.419cm}
    
    \begin{subfigure}[b]{0.66\textwidth}
        \includegraphics[width=\textwidth]{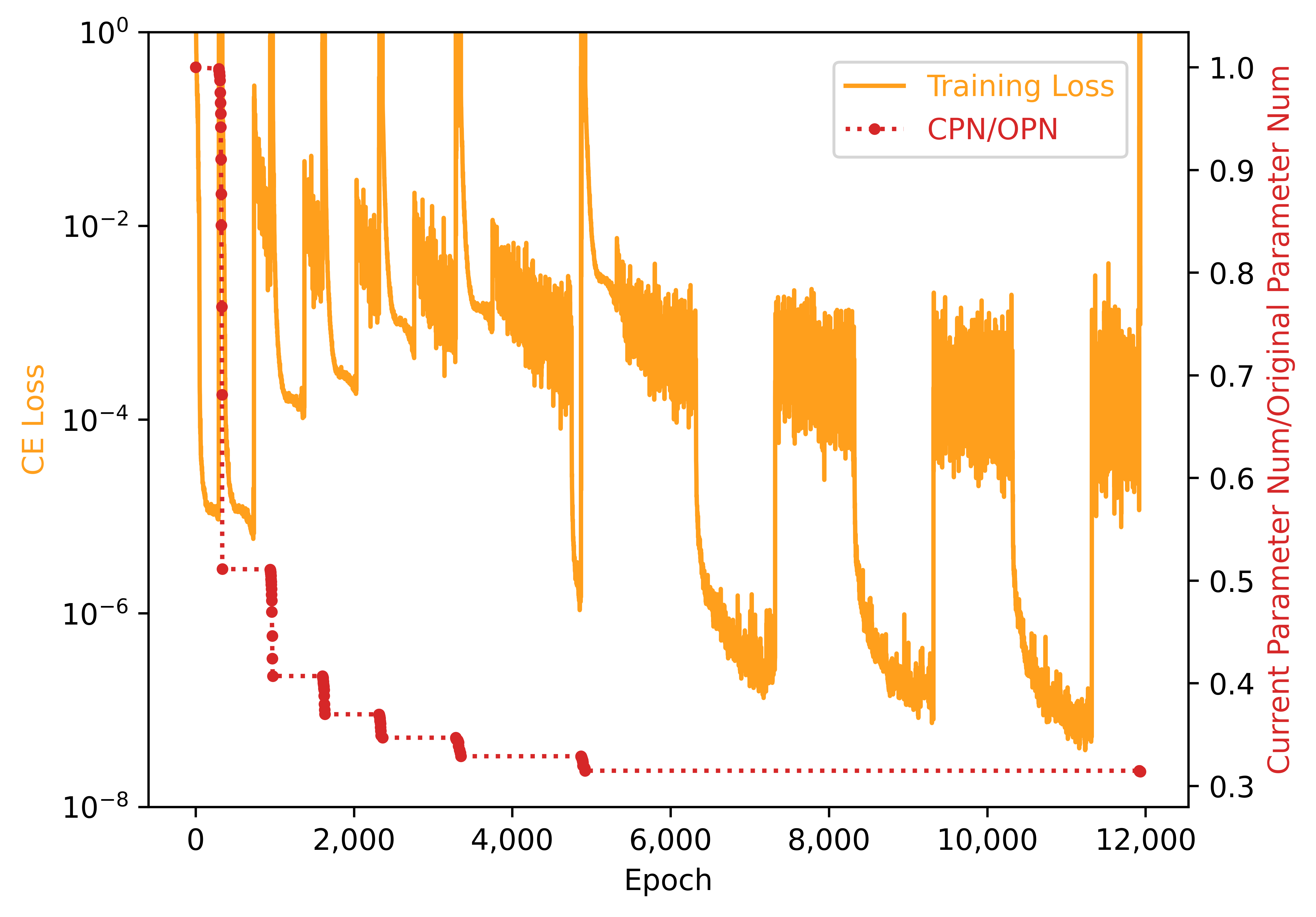}
        \caption{Training Loss.}
        \label{Conv_Training_Loss}
    \end{subfigure}

    \caption{Training loss and accuracy with parameter nums of CNN model. CPN stands for Current Parameter Number, OPN stands for Original Parameter Number, and CPN/OPN represents their ratio. 
    Each red dot in the figure represents a single-layer reduction. 
    Subfigures (\textbf{a}) and (\textbf{b}), respectively, show the relationship between model parameter nums and accuracy and training error. It is clearly observable that as the number of parameters decreases, accuracy tends to decrease and training error tends to increase. However, both metrics quickly return to their original levels.}
    \label{fig:cnnaccloss}
\end{figure}

\begin{figure}[H]
\includegraphics[width=13cm]{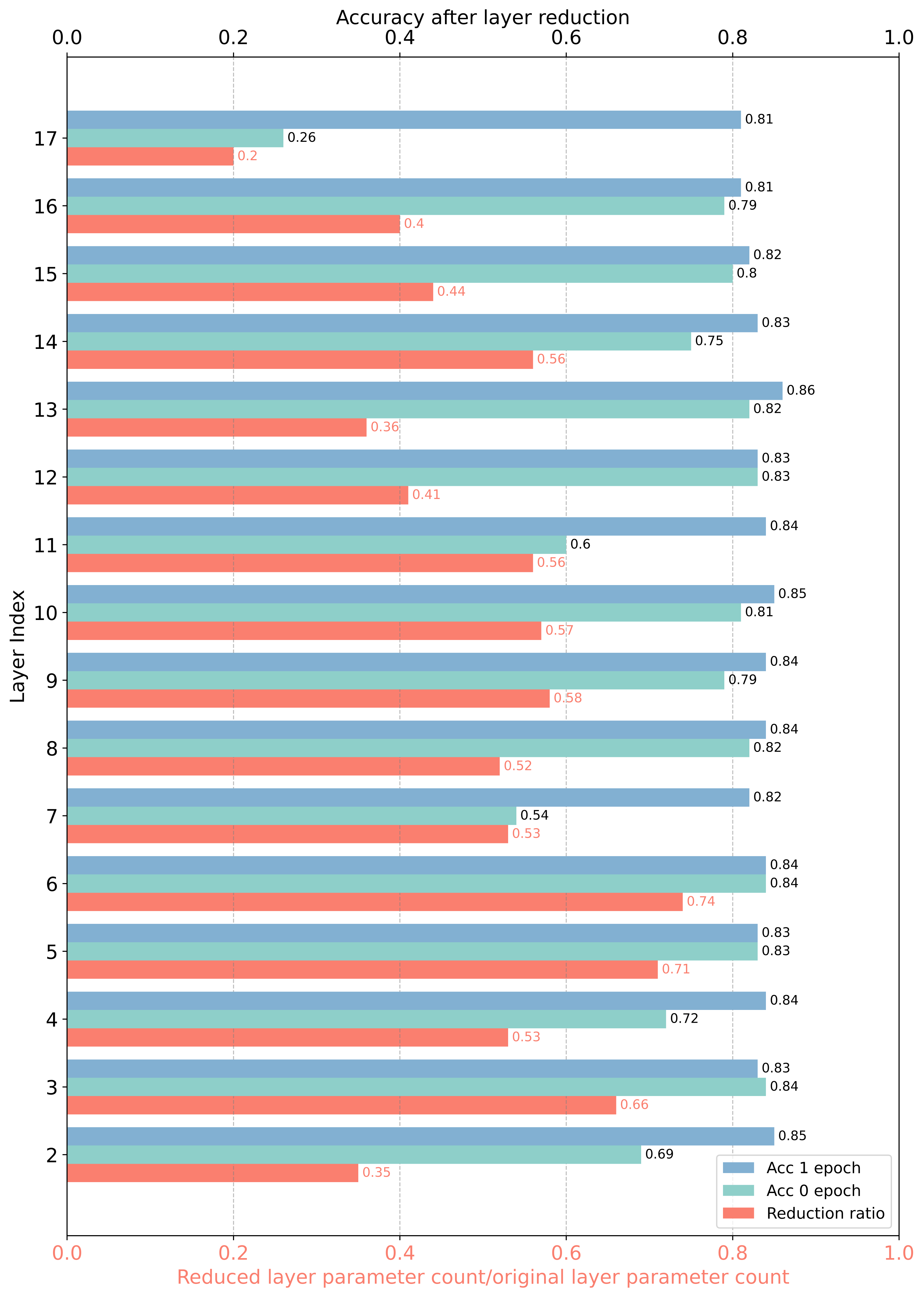}
\caption{First time of main redcution. The red reduction ratio indicates the proportion of the layer's parameters after reduction relative to the original number of parameters; a smaller ratio suggests a greater reduction. The green ``Acc 0 epoch'' represents the accuracy of the network after reduction without any further training, while the blue ``Acc 1 epoch'' represents the accuracy of the network after reduction with only one step of training.}
\label{Conv_1stReduction}
\end{figure}

In the CIFAR10 image classification task, we can quickly measure the performance of neural networks using the accuracy of the validation set, thus, we adopt automatic condensation reduction. 
Please refer to Table~\ref{Comparison_CNN} for a comparison of the reduction processes.  
The total condensation reduction training process is shown in the figure~\ref{Conv_Acc} and figure~\ref{Conv_Training_Loss}. From the start of training to the beginning of the first major reduction, it took 293 steps, at which point the model reached a validation set accuracy of 88.16\%. The major reduction was completed in 338 steps, reducing from the 2nd layer to the 17th layer, totaling 16 reductions. Initially, the main reduction criterion was set at 0.88 and the layer-by-layer reduction criterion at 0.84; by the end, these were adjusted to 0.87837 and 0.83837, respectively. During this major reduction, the condensation threshold for each layer was maintained at the original value of 0.88080. The original model's parameter count was reduced from 2,236,682 to 1,143,421, meaning the reduced model's parameters accounted for only 51.1\% of the original model. As shown in the figure~\ref{Conv_1stReduction}, throughout the major reduction process, most layers underwent significant reductions, with reduction ratios around 50\%. The greatest reduction was in the 17th layer, with a reduction ratio of 19.69\% and a single-layer parameter reduction of 380,103. Generally, the model exhibited minor performance deviations after each layer's reduction, with post-reduction, untrained validation set accuracies mostly above 70\%. After significant reductions, accuracy could reach over 80\% with just one step of training per layer. The observed reduction in parameter count validates the prevalence of condensation phenomena in convolutional neural networks, with a significant number of neurons condensing around a condensation threshold of approximately 0.88. The minor shifts resulting from reductions confirm that the condensation reduction method extended from fully connected to convolutional neural networks is reasonable and effective. The ability to rapidly restore model performance post-reduction suggests that the critical points achieved by the original model likely coincide with those of the reduced subnetworks. At step 385, 47 steps post-reduction, the reduced model's validation set accuracy exceeded 87\%. Considering model deployment objectives, it's only necessary to proceed with an additional 45 steps of reduction training after meeting deployment requirements to reduce nearly half of the parameter count. An additional 47 steps of training can yield a reduced model with negligible performance differences from the original, offering substantial practical value.

\begin{table}[H]
\caption{
 Comparison table of reduction process of CNN. "Parameter" represents the number of parameters in the neural network. Reduction rate represents the ratio of the current network parameter count to the original network parameter count. A lower reduction rate indicates a smaller network relative to the original, demonstrating a more effective reduction. Accuracy refers to the accuracy on the test set. From the table, it is clear that even with significant changes in the reduction rate, the accuracy only fluctuates slightly, showcasing the coalescence reduction algorithm's effectiveness in reducing the network size while maintaining model performance.}
\label{Comparison_CNN}
\begin{tabular}{|l|l|l|l|}
\hline
Model         & Parameter & Reduction ratio & Accuracy \\ \hline
Original      & 2,236,682 & 100\%           & 88.16\%  \\ \hline
1st reduced   & 1,143,421 & 51.12\%         & 88.01\%  \\ \hline
2nd reduced   & 910,370   & 40.70\%         & 88.00\%  \\ \hline
3rd reduced   & 827,032   & 36.98\%         & 87.24\%  \\ \hline
4th reduced   & 776,011   & 34.69\%         & 86.36\%  \\ \hline
5th reduced   & 735,527   & 32.88\%         & 85.88\%  \\ \hline
6th reduced   & 704,004   & 31.48\%         & 85.80\%  \\ \hline
7th reduced   & 669,228   & 29.92\%         & 83.88\%  \\ \hline
Final reduced & 258,212   & 11.54\%         & 83.21\%  \\ \hline
\end{tabular}
\end{table}

\unskip

The second main reduction began at step 942, with the model's validation set accuracy reaching 0.8801, and was completed by step 975. Initially, the main reduction criterion was set at 0.87837 and the layer-by-layer reduction criterion at 0.83837; by the end, these were adjusted to 0.87343 and 0.83343, respectively. Throughout this major reduction process, the condensation threshold for each layer remained at the original value of 0.88080, with no instances of reduction failure necessitating a rollback. The model's parameter count was further reduced from 1,143,421 (after the first reduction) to 910,370, with the new reduced model's parameters representing only 40.7\% of the original model. After the first reduction, the condensation level of the neural network's layers decreased, which was expected. According to the embedding principle, reduction starts from a specific critical point of the original model, which should correspond to an critical point of a certain fixed sub-network. The reduction process progressively approximates this sub-network by merging condensed neurons. As reduction proceeds, the reduced model becomes increasingly similar to the sub-network, resulting in fewer condensed neurons and a decrease in the neural network's redundant complexity, making further reductions increasingly challenging. When applying condensation reduction to practical problems, the reduction process can be prematurely concluded based on specific needs. In this study, to explore the limits of condensation reduction, we continued the reduction process as far as practically possible. Similar to the first reduction, each layer's reduction resulted in minimal performance deviation of the neural network, with rapid recovery of performance after just one step of training.

Our experiments continued until the completion of the sixth major reduction, with the third to sixth major reductions starting at steps 1601, 2316, 3286, 4867, 11919, and ending at steps 1633, 2362, 3351, 4921, 11919, respectively. The initial main reduction criteria at the start were 0.87343, 0.86641, 0.85904, 0.85315, and 0.85018, while the layer-by-layer reduction criteria were 0.83343, 0.82641, 0.81904, 0.81315, and 0.81018, respectively. The parameter counts of the reduced models were 827,032, 776,011, 735,527, 704,004, and 669,228, respectively; after the sixth major reduction, the reduced model's parameters accounted for only 29.9\% of the original model. It can be observed that each subsequent reduction became increasingly difficult, with longer training steps required, aligning with the expectation of approximating the ideal sub-network. Each layer's reduction still maintained the model's performance relatively unchanged. By the time of the sixth reduction's completion, there had been a total of 96 successful reductions, with a parameter decrease of 1,567,454, but only a slight decrease in the model's validation set accuracy, achieving 83.13\%, which is not significantly lower than the historical highest validation set accuracy of 88.16\%.

The final layer reduction, starting at step 15,200, utilized a condensation threshold set at 0.4, resulting in the model's parameter count being reduced to 258,212.
This is merely 11.5\% of the original model's parameters, and it is considered a very small model for tasks on CIFAR10. The reduction of the final layer alone eliminated 411,016 parameters, cutting over 60\% of the model's parameters following the sixth major reduction. Despite this significant reduction, the accuracy rate on the validation set without further training still reached 75.45\%. After training, the final reduced model achieved a peak accuracy of 83.21\%, virtually unchanged from before the reduction. This serves as yet another powerful validation of the effectiveness of condensation reduction.

\section{Discussion}

The successful application of condensation reduction in both fully connected and convolutional networks for practical problems demonstrates the universality of the condensation phenomenon. In the complex task of combustion simulation, condensation reduction effectively reduces network size while maintaining the accuracy of turbulence ignition predictions, confirming that a condensed network and its sub-networks are approximately equivalent. In image classification tasks, particularly in the broader application within depthwise separable convolutional neural networks, the definition and scope of condensation have been expanded. This extension enables the broader application of condensation reduction across various network types and tasks.

\appendix
\section[\appendixname~\thesection]{Appendix}
\subsection[\appendixname~\thesubsection]{Detail of condensation in CNN}
\label{A_condensation_cnn}
Consider a neural network consisting of two convolutional layers and one fully connected layer, with a structure of
\begin{equation}
d_{in}-m_{C}^{[1]}-m_{C}^{[2]}-m_{F_1}-d_{out}.
\end{equation}
Input as
\begin{equation}
\text{Input}_1 = (x_n)^{N_1\times (C_{inchannel}^{[0]} \times \text{Size}^{[1]})},
~~~~~~n\in (1,2,\cdots,N_1).
\end{equation}
Here $N_1$ represents the number of moves of the first convolutional layer in the input.
After passing through the first convolutional layer, the intermediate output obtained is
\begin{equation}
\text{Output}_1^{N_1\times C_{outchannel}^{[1]}} = \begin{bmatrix}
\vec{y}^{C_{outchannel}^{[1]}}_1\\
\vec{y}^{C_{outchannel}^{[1]}}_2\\
\vdots\\
\vec{y}^{C_{outchannel}^{[1]}}_{N_1}\\
\end{bmatrix}.
\end{equation}
The input data processed by each movement of the second convolutional kernel is
\begin{equation}
\text{input}_2^{C_{outchannel}^{[1]}\times  \text{Size}^{[2]}} =\begin{bmatrix}
\vec{y}^{C_{outchannel}^{[1]}}_1\\
\vec{y}^{C_{outchannel}^{[1]}}_2\\
\vdots\\
\vec{y}^{C_{outchannel}^{[1]}}_{\text{Size}^{[2]}}\\
\end{bmatrix}^{\boldsymbol{\top}}.
\end{equation}
The output obtained from each movement is
\begin{equation}
\text{output}_2^{C_{outchannel}^{[2]}} =\sigma(W_{[2]}^{C_{outchannel}^{[2]}\times (C_{outchannel}^{[1]}\times \text{Size}^{[2]})} \cdot \text{intput}^{C_{outchannel}^{[1]}\times \text{Size}^{[2]}}).
\end{equation}
If the direction of the $p$-th neuron and $q$-th neuron in the first convolutional layer is consistent, i.e. when the cosine similarity is $1$, there is
\begin{equation}
\boldsymbol{\vec{v}}_q^{\text{Size}^{[1]}\times C_{inchannel}^{[1]}} = \lambda \boldsymbol{\vec{v}}_p^{\text{Size}^{[1]}\times C_{inchannel}^{[1]}}, ~\lambda \in \mathbb{R}^+.
\end{equation}
By utilizing the homogeneity of the activation function, there are
\begin{equation}
\begin{aligned}
\vec{y}^{C_{outchannel}^{[1]}} &= \sigma(W_{[1]}^{C_{outchannel}^{[1]}\times (C_{inchannel}^{[1]}\times \text{Size}^{[1]})} \cdot \vec{x}^{C_{inchannel}^{[1]}\times \text{Size}^{[1]}})^{\boldsymbol{\top}}\\
&=\begin{bmatrix}
\vdots\\
\sigma(\boldsymbol{\vec{v}}_p \cdot \vec{x})\\
\vdots\\
\sigma(\boldsymbol{\vec{v}}_q \cdot \vec{x})\\
\vdots\\
\end{bmatrix}^{\boldsymbol{\top}}
=
\begin{bmatrix}
\vdots\\
\sigma(\boldsymbol{\vec{v}}_p \cdot \vec{x})\\
\vdots\\
\lambda\sigma(\boldsymbol{\vec{v}}_p \cdot \vec{x})\\
\vdots\\
\end{bmatrix}^{\boldsymbol{\top}}.
\end{aligned}
\end{equation}
For each convolutional kernel movement, it holds, so the total output of the first convolutional layer can be represented as
\begin{equation}
\text{Output}_1^{N_1\times C_{outchannel}^{[1]}} = \begin{bmatrix}
\vec{y}^{C_{outchannel}^{[1]}}_1\\
\vec{y}^{C_{outchannel}^{[1]}}_2\\
\vdots\\
\vec{y}^{C_{outchannel}^{[1]}}_{N_1}\\
\end{bmatrix}
=\begin{bmatrix}
\cdots &y^p_1 &\cdots &\lambda y^p_1 &\cdots\\
\cdots &y^p_2 &\cdots &\lambda y^p_2 &\cdots\\
\vdots &\vdots  &\vdots &\vdots &\vdots\\
\cdots &y^p_{N_1} &\cdots &\lambda y^p_{N_1} &\cdots\\
\end{bmatrix}.
\end{equation}
So the output obtained from each movement of the second convolutional layer is
\begin{equation}
\begin{aligned}
\text{output}_2^{C_{outchannel}^{[2]}} &=\sigma(W_{[2]}^{C_{outchannel}^{[2]}\times (C_{outchannel}^{[1]}\times \text{Size}^{[2]})} \cdot \text{intput}^{C_{outchannel}^{[1]}\times \text{Size}^{[2]}})\\
\end{aligned}
\end{equation}
Among them, the specific output of each output dimension is
\begin{equation}
\begin{aligned}
out &= \sigma(\boldsymbol{\vec{v}}\cdot\text{intput}^{C_{outchannel}^{[1]}\times \text{Size}^{[2]}})\\
&= \sigma(\sum_i^{C_{outchannel}^{[1]}}\sum_j^{\text{Size}^{[2]}}v_{ij}\cdot \text{intput}_{ij})\\
&= \sigma(\sum_{i\neq p,q}^{C_{outchannel}^{[1]}}\sum_j^{\text{Size}^{[2]}}v_{ij}\cdot \text{intput}_{ij}
+\sum_j^{\text{Size}^{[2]}}v_{qj}\cdot \text{intput}_{pj}
+\lambda\sum_j^{\text{Size}^{[2]}}v_{pj}\cdot \text{intput}_{pj})\\
&= \sigma(\sum_{i\neq p,q}^{C_{outchannel}^{[1]}}\sum_j^{\text{Size}^{[2]}}v_{ij}\cdot \text{intput}_{ij}
+\sum_j^{\text{Size}^{[2]}}(v_{pj}+\lambda v_{qj})\cdot \text{intput}_{pj}).\\
\end{aligned}
\end{equation}
Thus, for the overall output of the second convolutional layer, merging $p$-th neuron and $q$-th neuron from the first convolutional layer requires only that the corresponding input dimensions and parameters of the second convolutional layer be appropriately adjusted. Then, the output of the neural network before and after the merge will be completely identical.
If two neurons with consistent directions appear in the second convolutional layer, then, similar to the previous derivation, it can be proven that merging these two directionally consistent neurons will not change the output of the neural network before and after the merger.

\subsection[\appendixname~\thesubsection]{Details of condensation reduction in traditional CNN}
\label{A_detials_reduc_cnn}
Let's assume that the first $N$ neurons are grouped together; then, the new parameters of the model after reduction are as follows:
\begin{equation}
\boldsymbol{\vec{u}}_{main}^{[new]} = \sum_k^{N}\frac{\|\boldsymbol{\vec{v}}_{k}\|_2}{\|\boldsymbol{\vec{v}}_{main}\|_2}\boldsymbol{\vec{u}}_{k},
~~~~~~~\boldsymbol{\vec{v}}_{main}^{[new]}=\boldsymbol{\vec{v}}_{main},
\end{equation}
wherein,
\begin{equation}
\begin{aligned}
W_{[l]}^{C_{outchannel}^{[l]}\times (C_{inchannel}^{[l]}\times \text{Size}^{[l]})} &= \begin{bmatrix}
\boldsymbol{\vec{v}}_1 \\
\vdots \\
\boldsymbol{\vec{v}}_N \\
\boldsymbol{\vec{v}}_{N+1}\\
\vdots \\
\boldsymbol{\vec{v}}_{C_{outchannel}^{[l]}}
\end{bmatrix},\\
~~~~~~~R^{C_{outchannel}^{[l]}\times(C_{outchannel}^{[l+1]}\times \text{Size}^{[l+1]})}&=\text{Reshape}(W_{[l+1]}^{C_{outchannel}^{[l+1]}\times (C_{outchannel}^{[l]}\times \text{Size}^{[l+1]})}) \\
&= \begin{bmatrix}
\boldsymbol{\vec{u}}_1 \\
\vdots \\
\boldsymbol{\vec{u}}_N \\
\boldsymbol{\vec{u}}_{N+1}\\
\vdots \\
\boldsymbol{\vec{u}}_{C_{outchannel}^{[l]}}
\end{bmatrix}.
\end{aligned}
\end{equation}
The new parameter matrix of the neural network obtained is
\begin{equation}
\begin{aligned}
W_{[l]_{new}}^{(C_{outchannel}^{[l]}-N+1)\times (C_{inchannel}^{[l]}\times \text{Size}^{[l]})} &= \begin{bmatrix}
\boldsymbol{\vec{v}}_{main}^{[new]}\\
\boldsymbol{\vec{v}}_{N+1}\\
\vdots \\
\boldsymbol{\vec{v}}_{C_{outchannel}^{[l]}}
\end{bmatrix},\\
~~~~~~~R_{new}^{(C_{outchannel}^{[l]}-N+1)\times(C_{outchannel}^{[l+1]}\times \text{Size}^{[l+1]})}&=\text{Reshape}(W_{[l+1]_{new}}^{C_{outchannel}^{[l+1]}\times ((C_{outchannel}^{[l]}-N+1)\times \text{Size}^{[l+1]})})\\
&= \begin{bmatrix}

\boldsymbol{\vec{u}}_{main}^{[new]} \\
\boldsymbol{\vec{u}}_{N+1}\\
\vdots \\
\boldsymbol{\vec{u}}_{C_{outchannel}^{[l]}}
\end{bmatrix}.
\end{aligned}
\end{equation}
By using the new neural network parameter matrix to create a new neural network, the condensation-based reduction can be completed. The above generalization still applies completely even after incorporating pooling layers.

\subsection[\appendixname~\thesubsection]{Details of condensation reduction in MobileNetV2}
\label{A_detials_reduc_dwcnn}
Let's proceed with the derivation for the scenario where a convolutional kernel is applied only once, and initially, let's not consider batch normalization. Assuming the input is as follows:
\begin{equation}
\text{Input} =\begin{bmatrix}

\boldsymbol{\vec{x}}_1 \\
\cdots\\
\boldsymbol{\vec{x}}_{C_{inchannel}}

\end{bmatrix},
~~~~~~\vec{x}_i = \begin{pmatrix}

x_{i1} &
\cdots &
x_{i\text{Size}}

\end{pmatrix}.
\end{equation}
Assuming that two neurons from a depthwise layer have condensed in the same direction, let's designate them as:
\begin{equation}
\boldsymbol{\vec{v}}_q^{\text{Size}} = \lambda \boldsymbol{\vec{v}}_p^{\text{Size}}, ~\lambda \in \mathbb{R}^+.
\end{equation}
Their corresponding neurons in the preceding pointwise layer are:
\begin{equation}
\vec{\alpha}^{C_{inchannel}}_p =
\begin{pmatrix}
a_{p}^{[1]} &
\cdots &
a_{p}^{[C_{inchannel}]}
\end{pmatrix},
~~~~~~
\vec{\alpha}^{C_{inchannel}}_q =
\begin{pmatrix}
a_{q}^{[1]} &
\cdots &
a_{q}^{[C_{inchannel}]}
\end{pmatrix},
\end{equation}
An arbitrary neuron in the subsequent pointwise layer is represented as:
\begin{equation}
\vec{\beta} = 
\begin{pmatrix}
b_{1} &
\cdots &
b_p &
\cdots &
b_q &
\cdots &
b_{C_{inchannel}}
\end{pmatrix}.
\end{equation}
At this point, related to these two condensed neurons, the output of this complete module would be:
\begin{equation}
\begin{aligned}
\text{Output} &= b_py_p+b_qy_q\\
y_p &= \boldsymbol{\vec{v}}_p^{\text{Size}}\cdot(\vec{\alpha}_p^{C_{inchannel}}\cdot\text{Input}^{C_{inchannel}\times\text{Size}})^{\boldsymbol{\top}}\\
y_q &= \boldsymbol{\vec{v}}_q^{\text{Size}}\cdot(\vec{\alpha}_q^{C_{inchannel}}\cdot\text{Input}^{C_{inchannel}\times\text{Size}})^{\boldsymbol{\top}}\\
&=\lambda\boldsymbol{\vec{v}}_p^{\text{Size}}\cdot(\vec{\alpha}_q^{C_{inchannel}}\cdot\text{Input}^{C_{inchannel}\times\text{Size}})^{\boldsymbol{\top}}.\\
\end{aligned}
\end{equation}
When the two neurons from the depthwise layer are merged, the output of this complete module at that moment would be:
\begin{equation}
\begin{aligned}
\text{Output}_{[new]} &= b_{p_{[new]}}y_{p_{[new]}}\\
y_{p_{[new]}} &= \boldsymbol{\vec{v}}_p^{\text{Size}}\cdot(\vec{\alpha}_{p_{[new]}}^{C_{inchannel}}\cdot\text{Input}^{C_{inchannel}\times\text{Size}})^{\boldsymbol{\top}}.\\
\end{aligned}
\end{equation}
Following the previous logic of condensation reduction, there should be:
\begin{equation}
b_{p_{[new]}} = b_p + \lambda b_q.
\end{equation}
Therefore, to ensure that the module's output remains consistent before and after the merge, it should be:
\begin{equation}
\begin{aligned}
\text{Output} &= b_p\boldsymbol{\vec{v}}_p^{\text{Size}}\cdot(\vec{\alpha}_p^{C_{inchannel}}\cdot\text{Input}^{C_{inchannel}\times\text{Size}})^{\boldsymbol{\top}}+b_q\boldsymbol{\vec{v}}_q^{\text{Size}}\cdot(\vec{\alpha}_q^{C_{inchannel}}\cdot\text{Input}^{C_{inchannel}\times\text{Size}})^{\boldsymbol{\top}}\\
&= b_p\boldsymbol{\vec{v}}_p^{\text{Size}}\cdot(\vec{\alpha}_p^{C_{inchannel}}\cdot\text{Input}^{C_{inchannel}\times\text{Size}})^{\boldsymbol{\top}}+\lambda b_q\boldsymbol{\vec{v}}_p^{\text{Size}}\cdot(\vec{\alpha}_q^{C_{inchannel}}\cdot\text{Input}^{C_{inchannel}\times\text{Size}})^{\boldsymbol{\top}}\\
&=b_{p_{[new]}}\boldsymbol{\vec{v}}_p^{\text{Size}}\cdot(\vec{\alpha}_{p_{[new]}}^{C_{inchannel}}\cdot\text{Input}^{C_{inchannel}\times\text{Size}})^{\boldsymbol{\top}}.\\
\end{aligned}
\end{equation}
This leads to the formulation of a system of linear equations.
\begin{equation}
\begin{cases}
b_{p_{[new]}} &= b_p + \lambda b_q \\
b_{p_{[new]}}\vec{\alpha}_{p_{[new]}}^{C_{inchannel}} &= b_p\vec{\alpha}_p^{C_{inchannel}} + \lambda b_q\vec{\alpha}_q^{C_{inchannel}}
\end{cases}.
\end{equation}
By using the new parameters to construct a new module, the condensation reduction can be completed. The above derivation can easily be extended to situations where multiple neurons have condensed, resulting in:
\begin{equation}
\begin{cases}
b_{main_{[new]}} &= \sum_{k=1}^N\frac{\|\boldsymbol{\vec{v}}_{k}\|_2}{\|\boldsymbol{\vec{v}}_{main}\|_2}b_k\\
b_{main_{[new]}}\vec{\alpha}_{main_{[new]}}^{C_{inchannel}} &= \sum_{k=1}^N\frac{\|\boldsymbol{\vec{v}}_{k}\|_2}{\|\boldsymbol{\vec{v}}_{main}\|_2}b_k\vec{\alpha}_k^{C_{inchannel}}    
\end{cases}.
\end{equation}
It's important to note that the system of linear equations should hold true for all values of $b$, making it an overdetermined system of linear equations. In practice, it is necessary to use the least squares method to solve it.

Now, we will extend our results to modules that incorporate batch normalization, continuing to consider the case where two neurons in the depthwise layer have condensed. It's important to highlight that all derivations remain valid even after the addition of the ReLU6 activation function. Let's continue to assume:
\begin{equation}
\boldsymbol{\vec{v}}_q^{\text{Size}} = \lambda \boldsymbol{\vec{v}}_p^{\text{Size}}, ~\lambda \in \mathbb{R}^+.
\end{equation}
The expression for batch normalization is:
\begin{equation}
BN(\text{tmp}) = \frac{\text{tmp}-\text{Mean}}{\sqrt{\text{Var}}}\times \gamma+\eta.
\end{equation}
We can simplify the expression for batch normalization. For the batch normalization following the preceding pointwise layer, let's set:
\begin{equation}
\begin{aligned}
f_p(\text{tmp}) &= \varepsilon_p\cdot\text{tmp} +\zeta_p,\\
f_q(\text{tmp}) &= \varepsilon_q\cdot\text{tmp} +\zeta_q,\\
\varepsilon_p &= \frac{\gamma_p}{\sqrt{\text{Var}_p}},\\
\zeta_p &= -\frac{\text{mean}_p}{\sqrt{\text{Var}_p}}+\eta_p.
\end{aligned}
\end{equation}
For the batch normalization following the depthwise layer, let's similarly set:
\begin{equation}
\begin{aligned}
g_p(\text{tmp}) &= \mu_p\cdot\text{tmp} +\nu_p,\\
g_q(\text{tmp}) &= \mu_q\cdot\text{tmp} +\nu_q,\\
\mu_p &= \frac{\gamma_p}{\sqrt{\text{Var}_p}},\\
\nu_p &= -\frac{\text{mean}_p}{\sqrt{\text{Var}_p}}+\eta_p.
\end{aligned}
\end{equation}
After merging the two depthwise neurons, first, following the condensation reduction method used for fully connected neural networks, ensure that the intermediate output before the batch normalization layer, which follows the depthwise layer, remains consistent with the previous output. Therefore, we have:
\begin{equation}
\begin{aligned}
\text{output}
= &\left [\varepsilon_{p}\boldsymbol{\vec{v}}_p^{\text{Size}}\cdot (\vec{\alpha}_{p}^{C_{inchannel}}\cdot\text{Input}^{C_{inchannel}\times\text{Size}})^{\boldsymbol{\top}}+\boldsymbol{\vec{v}}_p^{\text{Size}}\cdot\zeta_{p}^{\boldsymbol{\top}}\right ]
+\\
&\left [\varepsilon_{q}\lambda\boldsymbol{\vec{v}}_p^{\text{Size}}\cdot (\vec{\alpha}_{q}^{C_{inchannel}}\cdot\text{Input}^{C_{inchannel}\times\text{Size}})^{\boldsymbol{\top}}+\lambda\boldsymbol{\vec{v}}_p^{\text{Size}}\cdot\zeta_{q}^{\boldsymbol{\top}}\right ]\\
= &\varepsilon_{p_{[new]}}\boldsymbol{\vec{v}}_p^{\text{Size}}\cdot (\vec{\alpha}_{p_{[new]}}^{C_{inchannel}}\cdot\text{Input}^{C_{inchannel}\times\text{Size}})^{\boldsymbol{\top}}+\boldsymbol{\vec{v}}_p^{\text{Size}}\cdot\zeta_{p_{[new]}}^{\boldsymbol{\top}}.\\
\end{aligned}
\end{equation}
So as to obtain
\begin{equation}
\begin{cases}
\varepsilon_{p_{[new]}}\vec{\alpha}_{p_{[new]}}^{C_{inchannel}} &= \varepsilon_{p}\vec{\alpha}_{p}^{C_{inchannel}} +\lambda\varepsilon_{q}\vec{\alpha}_{q}^{C_{inchannel}}\\
\zeta_{p_{[new]}} &= \zeta_{p} + \lambda \zeta_{q}
\end{cases}.
\end{equation}
At this point, the output of this complete module, which is related to these two condensed neurons, is
\begin{equation}
\begin{aligned}
\text{Output} &= b_pg_p(y_p)+b_qg_q(y_q)\\
y_p &= \boldsymbol{\vec{v}}_p^{\text{Size}}\cdot f_p((\vec{\alpha}_p^{C_{inchannel}}\cdot\text{Input}^{C_{inchannel}\times\text{Size}})^{\boldsymbol{\top}})\\
y_q &= \boldsymbol{\vec{v}}_q^{\text{Size}}\cdot f_q((\vec{\alpha}_q^{C_{inchannel}}\cdot\text{Input}^{C_{inchannel}\times\text{Size}})^{\boldsymbol{\top}})\\
&=\lambda\boldsymbol{\vec{v}}_p^{\text{Size}}\cdot f_q((\vec{\alpha}_q^{C_{inchannel}}\cdot\text{Input}^{C_{inchannel}\times\text{Size}})^{\boldsymbol{\top}}).\\
\end{aligned}
\end{equation}
After merging two neurons, the output of the complete module is
\begin{equation}
\begin{aligned}
\text{Output}_{[new]} &= b_{p_{[new]}}g_{p_{[new]}}(y_{p_{[new]}})\\
y_{p_{[new]}} &= \boldsymbol{\vec{v}}_p^{\text{Size}}\cdot f_{p_{[new]}}((\vec{\alpha}_{p_{[new]}}^{C_{inchannel}}\cdot\text{Input}^{C_{inchannel}\times\text{Size}})^{\boldsymbol{\top}})\\
&=\boldsymbol{\vec{v}}_p^{\text{Size}}\cdot (\varepsilon_{p_{[new]}}(\vec{\alpha}_{p_{[new]}}^{C_{inchannel}}\cdot\text{Input}^{C_{inchannel}\times\text{Size}})+\zeta_{p_{[new]}})^{\boldsymbol{\top}}\\
& = \varepsilon_{p_{[new]}}\boldsymbol{\vec{v}}_p^{\text{Size}}\cdot (\vec{\alpha}_{p_{[new]}}^{C_{inchannel}}\cdot\text{Input}^{C_{inchannel}\times\text{Size}})^{\boldsymbol{\top}}+\boldsymbol{\vec{v}}_p^{\text{Size}}\cdot\zeta_{p_{[new]}}^{\boldsymbol{\top}}.
\end{aligned}
\end{equation}
Therefore, in order to ensure consistent output of the modules before and after merging, there should be
\begin{equation}
\begin{aligned}
\text{Output} 
 = &b_{p}g_{p}(\varepsilon_{p}\boldsymbol{\vec{v}}_p^{\text{Size}}\cdot (\vec{\alpha}_{p}^{C_{inchannel}}\cdot\text{Input}^{C_{inchannel}\times\text{Size}})^{\boldsymbol{\top}}+\boldsymbol{\vec{v}}_p^{\text{Size}}\cdot\zeta_{p}^{\boldsymbol{\top}})
+\\
&b_{q}g_{q}(\varepsilon_{q}\lambda\boldsymbol{\vec{v}}_p^{\text{Size}}\cdot (\vec{\alpha}_{q}^{C_{inchannel}}\cdot\text{Input}^{C_{inchannel}\times\text{Size}})^{\boldsymbol{\top}}+\lambda\boldsymbol{\vec{v}}_p^{\text{Size}}\cdot\zeta_{q}^{\boldsymbol{\top}})\\
 = &b_{p_{[new]}}g_{p_{[new]}}(\varepsilon_{p_{[new]}}\boldsymbol{\vec{v}}_p^{\text{Size}}\cdot (\vec{\alpha}_{p_{[new]}}^{C_{inchannel}}\cdot\text{Input}^{C_{inchannel}\times\text{Size}})^{\boldsymbol{\top}}+\boldsymbol{\vec{v}}_p^{\text{Size}}\cdot\zeta_{p_{[new]}}^{\boldsymbol{\top}}).
\end{aligned}
\end{equation}
Set
\begin{equation}
\begin{aligned}
\vec{x}_p &=\vec{\alpha}_{p}^{C_{inchannel}}\cdot\text{Input}^{C_{inchannel}\times\text{Size}},\\
\vec{x}_q &=\vec{\alpha}_{q}^{C_{inchannel}}\cdot\text{Input}^{C_{inchannel}\times\text{Size}},\\
\vec{x_{[new]}} &=\vec{\alpha}_{p_{[new]}}^{C_{inchannel}}\cdot\text{Input}^{C_{inchannel}\times\text{Size}}.\\
\end{aligned}
\end{equation}
In order to ensure consistent output of the modules before and after merging, there are
\begin{equation}
\begin{aligned}
\text{Output} 
 = &b_{p}g_{p}(\varepsilon_{p}\boldsymbol{\vec{v}}_p^{\text{Size}}\cdot \vec{x}_p^{\boldsymbol{\top}}+\boldsymbol{\vec{v}}_p^{\text{Size}}\cdot\zeta_{p}^{\boldsymbol{\top}})
+b_{q}g_{q}(\varepsilon_{q}\lambda\boldsymbol{\vec{v}}_p^{\text{Size}}\cdot \vec{x}_q^{\boldsymbol{\top}}+\lambda\boldsymbol{\vec{v}}_p^{\text{Size}}\cdot\zeta_{q}^{\boldsymbol{\top}})\\
 = &b_{p}\mu_p(\varepsilon_{p}\boldsymbol{\vec{v}}_p^{\text{Size}}\cdot \vec{x}_p^{\boldsymbol{\top}}+\boldsymbol{\vec{v}}_p^{\text{Size}}\cdot\zeta_{p}^{\boldsymbol{\top}})
+b_{q}\mu_q(\varepsilon_{q}\lambda\boldsymbol{\vec{v}}_p^{\text{Size}}\cdot \vec{x}_q^{\boldsymbol{\top}}+\lambda\boldsymbol{\vec{v}}_p^{\text{Size}}\cdot\zeta_{q}^{\boldsymbol{\top}})
+b_{p}\nu_{p} +b_{q}\nu_{q}\\
 = &(b_{p}\mu_p\varepsilon_{p}\boldsymbol{\vec{v}}_p^{\text{Size}}\cdot \vec{x}_p^{\boldsymbol{\top}}+ \lambda b_{q}\mu_q\varepsilon_{q}\boldsymbol{\vec{v}}_p^{\text{Size}}\cdot \vec{x}_q^{\boldsymbol{\top}}) +\\
&(b_{p}\mu_p\boldsymbol{\vec{v}}_p^{\text{Size}}\cdot\zeta_{p}^{\boldsymbol{\top}}+\lambda b_{q}\mu_q\boldsymbol{\vec{v}}_p^{\text{Size}}\cdot\zeta_{q}^{\boldsymbol{\top}})
+b_{p}\nu_{p} +b_{q}\nu_{q}\\
 = &b_{p_{[new]}}g_{p_{[new]}}(\varepsilon_{p_{[new]}}\boldsymbol{\vec{v}}_p^{\text{Size}}\cdot \vec{x_{[new]}}^{\boldsymbol{\top}}+\boldsymbol{\vec{v}}_p^{\text{Size}}\cdot\zeta_{p_{[new]}}^{\boldsymbol{\top}})\\
 = &b_{p_{[new]}}\mu_{p_{[new]}}\varepsilon_{p_{[new]}}\boldsymbol{\vec{v}}_p^{\text{Size}}\cdot \vec{x_{[new]}}^{\boldsymbol{\top}}+
b_{p_{[new]}}\mu_{p_{[new]}}\boldsymbol{\vec{v}}_p^{\text{Size}}\cdot\zeta_{p_{[new]}}^{\boldsymbol{\top}}
+b_{p_{[new]}}\nu_{p_{[new]}}.
\end{aligned}
\end{equation}
Combine the equations obtained from the intermediate output to obtain a system of linear equations
\begin{equation}
\begin{cases}
b_{p_{[new]}}\mu_{p_{[new]}}\varepsilon_{p_{[new]}}\boldsymbol{\vec{v}}_p^{\text{Size}}\cdot \vec{x_{[new]}}^{\boldsymbol{\top}} &= 
(b_{p}\mu_p\varepsilon_{p}+ \lambda b_{q}\mu_q\varepsilon_{q})\boldsymbol{\vec{v}}_p^{\text{Size}}\cdot \vec{x}^{\boldsymbol{\top}}\\
b_{p_{[new]}}\nu_{p_{[new]}} &= b_{p}\nu_{p} + b_{q}\nu_{q}\\
\varepsilon_{p_{[new]}}\vec{\alpha}_{p_{[new]}}^{C_{inchannel}} &= \varepsilon_{p}\vec{\alpha}_{p}^{C_{inchannel}} +\lambda\varepsilon_{q}\vec{\alpha}_{q}^{C_{inchannel}}\\
\zeta_{p_{[new]}} &= \zeta_{p} + \lambda \zeta_{q}
\end{cases}.
\end{equation}
So as to obtain
\begin{equation}
\begin{cases}
b_{p_{[new]}}\mu_{p_{[new]}}\varepsilon_{p_{[new]}} \vec{\alpha}_{p_{[new]}}^{C_{inchannel}} &= 
b_{p}\mu_p\varepsilon_{p}\vec{\alpha}_{p}^{C_{inchannel}}+ \lambda b_{q}\mu_q\varepsilon_{q}\vec{\alpha}_{q}^{C_{inchannel}}\\
b_{p_{[new]}}\nu_{p_{[new]}} &= b_{p}\nu_{p} + b_{q}\nu_{q}\\
\varepsilon_{p_{[new]}}\vec{\alpha}_{p_{[new]}}^{C_{inchannel}} &= \varepsilon_{p}\vec{\alpha}_{p}^{C_{inchannel}} +\lambda\varepsilon_{q}\vec{\alpha}_{q}^{C_{inchannel}}\\
\zeta_{p_{[new]}} &= \zeta_{p} + \lambda \zeta_{q}
\end{cases}.
\end{equation}
Both $b$ and $a$ in the equation are solved using the previously derived equation system without batch standardization. The total equation system is
\begin{equation}
\begin{cases}
b_{p_{[new]}}\mu_{p_{[new]}}\varepsilon_{p_{[new]}} \vec{\alpha}_{p_{[new]}}^{C_{inchannel}} &= 
b_{p}\mu_p\varepsilon_{p}\vec{\alpha}_{p}^{C_{inchannel}}+ \lambda b_{q}\mu_q\varepsilon_{q}\vec{\alpha}_{q}^{C_{inchannel}}\\
b_{p_{[new]}}\nu_{p_{[new]}} &= b_{p}\nu_{p} + b_{q}\nu_{q}\\
\varepsilon_{p_{[new]}}\vec{\alpha}_{p_{[new]}}^{C_{inchannel}} &= \varepsilon_{p}\vec{\alpha}_{p}^{C_{inchannel}} +\lambda\varepsilon_{q}\vec{\alpha}_{q}^{C_{inchannel}}\\
\zeta_{p_{[new]}} &= \zeta_{p} + \lambda \zeta_{q}\\
b_{p_{[new]}} &= b_p + \lambda b_q \\
b_{p_{[new]}}\vec{\alpha}_{p_{[new]}}^{C_{inchannel}} &= b_p\vec{\alpha}_p^{C_{inchannel}} + \lambda b_q\vec{\alpha}_q^{C_{inchannel}}
\end{cases}.
\end{equation}
The equation system that extends to the condensation of multiple neurons is
\begin{equation}
\begin{cases}
b_{main_{[new]}} &= \sum_{k=1}^N\frac{\|\boldsymbol{\vec{v}}_{k}\|_2}{\|\boldsymbol{\vec{v}}_{main}\|_2}b_k\\
b_{main_{[new]}}\vec{\alpha}_{main_{[new]}}^{C_{inchannel}} &= \sum_{k=1}^N\frac{\|\boldsymbol{\vec{v}}_{k}\|_2}{\|\boldsymbol{\vec{v}}_{main}\|_2}b_k\vec{\alpha}_k^{C_{inchannel}}\\
\varepsilon_{main_{[new]}}\vec{\alpha}_{main_{[new]}}^{C_{inchannel}} &= 
\sum_{k=0}^N\frac{\|\boldsymbol{\vec{v}}_{k}\|_2}{\|\boldsymbol{\vec{v}}_{main}\|_2}\varepsilon_{k}\vec{\alpha}_{k}^{C_{inchannel}}\\
\zeta_{main_{[new]}} &= 
\sum_{k=0}^N\frac{\|\boldsymbol{\vec{v}}_{k}\|_2}{\|\boldsymbol{\vec{v}}_{main}\|_2}\zeta_{k}\\
b_{main_{[new]}}\mu_{main_{[new]}}\varepsilon_{main_{[new]}} \vec{\alpha}_{main_{[new]}}^{C_{inchannel}} &= 
\sum_{k=0}^N\frac{\|\boldsymbol{\vec{v}}_{k}\|_2}{\|\boldsymbol{\vec{v}}_{main}\|_2}
b_{k}\mu_k\varepsilon_{k}\vec{\alpha}_{k}^{C_{inchannel}}\\
b_{main_{[new]}}\nu_{main_{[new]}} &=
\sum_{k=0}^N\frac{\|\boldsymbol{\vec{v}}_{k}\|_2}{\|\boldsymbol{\vec{v}}_{main}\|_2}
b_{k}\nu_k\\
\end{cases}.
\end{equation}
The pointwise layer's each neuron corresponds to one equation in this system, effectively constructing an overdetermined system of linear equations. We still utilize the method of least squares for solving.

Afterwards, calculate the mean and variance of batch standardization using the equation
\begin{equation}
\left\{
\begin{array}{c}

\text{Mean} = 
\sum_{k=0}^N(\frac{\|\boldsymbol{\vec{v}}_{k}\|_2}{\|\boldsymbol{\vec{v}}_{main}\|_2})^{-1}\text{Mean}_k\\

\text{Var} = 
\sum_{k=0}^N(\frac{\|\boldsymbol{\vec{v}}_{k}\|_2}{\|\boldsymbol{\vec{v}}_{main}\|_2})^{-2}\text{Var}_k

\end{array}
\right.
.
\end{equation}
Afterwards, the remaining parameters can be easily obtained, and a reduced model can be obtained using the new parameter matrix.

\bibliographystyle{alpha}
\bibliography{bibref.bib}

\end{document}